\documentclass[journal,10pt]{IEEEtran}   

\IEEEoverridecommandlockouts
\usepackage{graphicx}
\usepackage[font={small,it}]{caption}
\usepackage{subcaption,url}
\usepackage{mathtools}
\usepackage{amsmath,amssymb,amsfonts}
\usepackage{xcolor,etoolbox}
\usepackage[nospace,noadjust]{cite}
\usepackage{algorithm,multirow} 
\usepackage{tikz}
\usepackage{algpseudocode}
\newcommand{\PP}{\mathbb{P}} % probability
\def\Trace{\mathsf{Tr}} %Trace operator
\newcommand\copyrighttext{%
  \footnotesize \textcopyright 2023 IEEE. Personal use of this material is permitted.  Permission from IEEE must be obtained for all other uses, in any current or future media, including reprinting/republishing this material for advertising or promotional purposes, creating new collective works, for resale or redistribution to servers or lists, or reuse of any copyrighted component of this work in other works. This article has been accepted for publication in IEEE Transactions on Intelligent Vehicles.}
\newcommand\copyrightnotice{%
\begin{tikzpicture}[remember picture,overlay]
\node[anchor=south,yshift=7pt] at (current page.south) {\fbox{\parbox{\dimexpr\textwidth-\fboxsep-\fboxrule\relax}{\copyrighttext}}};
\end{tikzpicture}%
}

\begin{document}
\title{Edge-assisted ML-aided Uncertainty-aware\\ Vehicle Collision Avoidance at Urban Intersections}

\author{Dinesh Cyril Selvaraj, Christian Vitale,~\IEEEmembership{Member,~IEEE}, Tania Panayiotou, Panayiotis Kolios,\\
Carla Fabiana Chiasserini,~\IEEEmembership{Fellow Member,~IEEE}, and~Georgios Ellinas,~\IEEEmembership{Senior Member,~IEEE \vspace{-3mm}}

\thanks{This work was supported in part by the European Union's Horizon 2020 Research and Innovation Programme under Grant No.\,739551 (KIOS CoE-TEAMING), Grant No.\,101003439 (C-AVOID), and 
Grant No.\,101069688 (CONNECT), and in part by the Republic of Cyprus through the Deputy Ministry of Research, Innovation and Digital Policy. Views and opinions expressed are however those of the authors only and do not necessarily reflect those of the European Union. Neither the European Union nor the granting authority can be held responsible for them. Computational resources were provided by HPC@POLITO (http://www.hpc.polito.it).}

\thanks{Dinesh Cyril Selvaraj and Carla Fabiana Chiasserini are with CARS@Polito and Politecnico di Torino, Torino, Italy (e-mail:{\tt\small \{dinesh.selvaraj, carla.chiasserini\}@polito.it}). Christian Vitale, Tania Panayiotou, Panayiotis Kolios, and Georgios Ellinas are with the KIOS Research and Innovation Center of Excellence and the Department of Electrical and Computer Engineering, University of Cyprus, Nicosia, Cyprus. (e-mail:{\tt\small \{vitale.christian, panayiotou.tania, pkolios, gellinas\}@ucy.ac.cy})}}

\maketitle
\copyrightnotice
\begin{abstract}
Intersection crossing represents one of the most dangerous sections of the road infrastructure and Connected Vehicles (CVs) can serve as a revolutionary solution to the problem. In this work, we present a novel framework that detects preemptively collisions at urban crossroads, exploiting the Multi-access Edge Computing (MEC) platform of 5G networks. At the MEC, an Intersection Manager (IM) collects information from both vehicles and the road infrastructure to create a holistic view of the area of interest. Based on the historical data collected, the IM leverages the capabilities of an encoder-decoder recurrent neural network to predict, with high accuracy, the future vehicles' trajectories. As, however, accuracy is not a sufficient measure of how much we can trust a model, trajectory predictions are additionally associated with a measure of uncertainty towards confident collision forecasting and avoidance. Hence, contrary to any other approach in the state of the art, an uncertainty-aware collision prediction framework is developed that is shown to detect well in advance (and with high reliability) if two vehicles are on a collision course. Subsequently, collision detection triggers a number of alarms that signal the colliding vehicles to brake. Under real-world settings, thanks to the preemptive capabilities of the proposed approach, all the simulated imminent dangers are averted.
\end{abstract}


\section{Introduction}
\label{sec:intro}

Recent studies have shown that accidents on the road infrastructure result in more than $1.35$ million worldwide casualties annually \cite{world2018global}, highlighting that safety is still of the utmost concern in the automotive sector. As a possible solution, new intelligent functionalities have been recently considered for the new generation of connected (and automated) vehicles that are about to enter the market \cite{bansal2017forecasting}. Specifically, several efforts exploit the fact that these vehicles can exchange information and create a better understanding of their surroundings, which, in turn, can help to improve driving assistance systems and avoid dangerous situations. As an example, an enriched view of the driving environment helps in refining the vehicles' trajectory predictions, which could be later used as the main tool to foresee possible collisions on the road infrastructure. 

As a consequence, predicting the trajectories of Connected Vehicles (CVs) is a well-researched topic. Nevertheless, predicting future collisions starting from trajectory predictions has been scarcely investigated, and still remains a challenging task. For a successful collision detection framework, indeed, trajectory predictions must be compared effectively, and, for any possible real-time safety application exploiting such a framework, in a time-efficient manner. Hence, \textit{when it comes to collision avoidance}, simplified trajectory predictions have been proposed, namely: (i) constant speed location projections \cite{Gomez_SCD}; (ii) Kalman Filter (KF) predictions, with a Time-To-Collision (TTC) metric being used as a risk factor \cite{Wang_TPCD}; and (iii) collision time estimation via polynomial approximations \cite{marco_CA}. However, the aforementioned approaches do not achieve the level of precision required for reliable and preemptive collision detection when a complex driving environment is considered, i.e., when human-driven vehicles at urban intersections are considered \cite{Dinesh21}. Indeed, due to the human factor, these simplified trajectory predictions cannot cope with the higher levels of uncertainty (i.e., input uncertainty is present) and with the more evident non-linearities. 

This work aims at filling this gap by presenting a framework that preemptively detects collisions, with short execution time, exploiting a precise model able to capture the highly non-linear and uncertain behavior of drivers, even at dense urban crossroads. Our framework is based on the ability of the Multi-access Edge Computing (MEC) platform of 5G networks to collect, thanks to both Vehicle-to-Infrastructure (V2I) and Infrastructure-to-Infrastructure (I2I) communications, relevant and extensive data sampled in real-time (e.g., regarding vehicle location and speed, but also traffic light phases) at an edge entity, named the Intersection Manager (IM). Such data is processed directly at the edge to obtain not only accurate trajectory predictions, but also an estimation of the associated prediction uncertainty. It is worth mentioning that, even if a few trajectory uncertainty estimation models exist in the literature, e.g., \cite{Huang_TPUncert}, these are obtained in a computationally expensive way, which is not appropriate for real-time decision-making. In our work, instead, trajectory uncertainty estimation is obtained through an estimation of the prediction intervals~\cite{Meeker17}, which capture indirectly the various, possible, modes of future vehicle trajectories without the need of defining the set of possible driver actions at the intersection.

More in detail, trajectory predictions and uncertainty estimations are obtained through the use of two lightweight independent encoder-decoder regression architectures with recurrent units, specifically designed for sequence-to-sequence prediction problems, i.e., using LSTM Encoder Decoder (LSTM-ED) models~\cite{cho2014learning}.  
Then, a Random Forest Classifier (RFC), working as an ensemble technique, is used to recognize patterns of trajectory predictions and uncertainty estimations that led to dangerous situations in the past, to timely determine if two vehicles are on  collision course. Thanks to the proposed approach, the IM can transmit alarms to the vehicles involved, so that the imminent danger is avoided.
The performance of the proposed approach is verified  through extensive tests performed on a diverse set of simulations drawn by real-world intersection data, with results of high quality in terms of: (i) detection rate; (ii) low false positive rate; and (iii) time between the alarm sent by the IM and the collision time, i.e., the available reaction time. In the obtained results, accounting for trajectory prediction uncertainty estimation allows for: (i) reducing the number of false collision detections when the point trajectory predictions suggest a collision but a large prediction uncertainty is associated with them; and (ii) improving the available reaction time to possible threats where a large uncertainty is associated with safe trajectory predictions. As a result, all dangerous situations are predicted largely in advance by the proposed framework, allowing to efficiently avoid the imminent collisions with simple maneuvers that can be applied also by human drivers, e.g., constant braking.

In the rest of the paper, Sec. \ref{sec:rel} reviews the state of the art on trajectory prediction and on collision avoidance algorithms at urban intersections, and summarizes the main contributions of this work (Sec. \ref{subsec:related_contribution}). Sec. \ref{sec:system_model} illustrates an overview on the proposed collision forecast framework, Sec. \ref{sec:LSTM_network} shows the trajectory prediction and the uncertainty estimation models, while Sec. \ref{sec:decision_tree} and Sec. \ref{sec:collision_avoidance} present the RFC technique that is used to forecast collisions, and how it can be used to avoid imminent dangers at urban intersections, respectively. Sec. \ref{sec:peva} showcases the obtained performance on a realistic real-world intersection data, including comparisons with a collision avoidance technique that only exploits point-based vehicle trajectory predictions and with the state-of-the-art approach in \cite{marco_CA}. Finally, Sec. \ref{sec:concl} concludes the work.

\section{Related Work\label{sec:rel}}
As a vital safety application in the automotive domain, diverse methodologies have been presented for preemptive vehicle collision detection \cite{CD_Survey}. As in our approach, in most of the proposed solutions, e.g., \cite{Tao_TPCD, Wang_TPCD, marco_CA, Ruifeng_TPCD}, a collision is detected thanks to the use of vehicle trajectory predictions. As such, we first discuss existing trajectory prediction approaches that can be applied to urban intersections (Sec. \ref{subsec:related_trajectory}), then we discuss previous works that propose collision avoidance frameworks (Sec. \ref{subsec:related_collision}), and finally, we summarize how our proposed approach advances the state of the art (Sec. \ref{subsec:related_contribution}).

\subsection{Vehicle Trajectory Prediction}
\label{subsec:related_trajectory}
The vehicle trajectory prediction problem is by itself a popular topic with significant research activity \cite{rudenko2020human}. Related works mostly focus on predicting vehicle trajectories for simple scenarios, e.g., in highway scenarios, or they exploit only a partial view of the road infrastructure (\cite{altche2017lstm, deo2018multi, kim2017probabilistic, jeong2020surround}) when more complex ones are analyzed.

This type of solution can hardly be extended to urban intersections. As an example, \cite{PersVTP} proposes a vehicle trajectory prediction framework where its output is based on a ``static'' categorization of the drivers' driving styles. Assuming that driving styles remain constant over time works well in a highway setting, but such an assumption does not hold at intersections, where the drivers' decisions depend on the specific scenario encountered. On the other hand, authors in \cite{Intention} use Gaussian Processes to define both the vehicle's expected trajectory and the uncertainty associated with the prediction, enumerating the probabilities associated with all possible driver intentions in a highway scenario. Considering the countless possible maneuvers at intersections, a similar approach is difficult to tune and does not scale.

Fewer works are present if the urban intersection use case is taken into account. A first result involves the prediction of drivers' maneuver intentions, i.e., predicting if a vehicle proceeds straight or turns at the intersection, rather than examining full trajectory predictions \cite{tran2014online, phillips2017generalizable}. More evolved approaches provide a ``group'' trajectory prediction framework, following the intuition that vehicles' interaction is critical to improve performance \cite{GI_MVTP,deo2018convolutional}. Similarly, single-vehicle trajectory predictions accounting for interaction-aware architectures based on Graphical Neural Networks (GNNs) have been proposed recently \cite{Gao, Liang}. Herein, both interactions among vehicles and between vehicles and the road infrastructure are modeled and accounted for to improve performance. Furthermore, a multimodal trajectory prediction framework generating multiple possible trajectories for each vehicle, and their corresponding likelihood, was proposed in \cite{Cui}. Even if the support of wireless communications has been envisioned in \cite{Cui}, and despite the fact that incorporating surrounding vehicle movements improves the prediction accuracy, the computation complexity increases multi-fold when all the possible maneuvers and interactions are modeled individually. Building on the same concept, hence suffering from the same limitations, \cite{TCN} makes use of Temporal Convolutional Networks and of Kalman Filter outputs, to also obtain a trajectory uncertainty estimation.

Alternatively, \cite{V2XTLTP} proposes an approach similar to the one presented in our work. A trajectory prediction model is obtained through the monitoring of a specific area by collecting information at a MEC server. A general trajectory prediction model is then transmitted back to the vehicles, to update the model based on their driving style. Even though interesting, this approach considers the area of interest divided into ``squares'', among which the movement of vehicles is predicted, hence suffering from the quantization effect. Furthermore, \cite{Huang_TPUncert} obtains a set of predictors based on different input sources (i.e., camera, odometry, and other sensor data). Herein, uncertainty is obtained as a confidence score and it is used to select, among the predictors obtained, the best trajectory prediction. Nevertheless, trajectory predictions based on a single source, even if the best at any point in time, are not as precise as building a trajectory prediction from multiple sources and properly accounting for the associated uncertainty.

Finally, it must be noted that none of the approaches presented above integrates the trajectory prediction framework with an approach that preemptively forecasts collisions among vehicles; hence, it is unclear if any of the advanced aforementioned approaches are able to fulfill the necessary critical time-sensitive requirements of a collision avoidance application.

\subsection{Collision Detection/Avoidance}
\label{subsec:related_collision}
The approaches present in the state of the art for collision detection/avoidance can be divided into two groups, i.e., in frameworks that forecast collisions based: (i) on information relative to the present state of the vehicles; and (ii) on information that involves predictions of future vehicles' states.

For the first class of collision prediction algorithms, two metrics are used to identify imminent dangers: (i) distance \cite{Gomez_SCD, Huang_SCD, Katare_SCD}, and (ii) time \cite{VCP, Ruifeng_TPCD, Wang_TPCD, Tao_TPCD}. In \cite{Gomez_SCD, Huang_SCD, Katare_SCD} distance-based collision detection methodologies utilize both real-time in-vehicle sensors’ measurements and/or information received through a communication framework. In \cite{VCP}, while similar input data are considered, a time-based approach is used to detect and avoid collisions. Specifically, authors in \cite{VCP} utilize a set of virtual collision points between any two lanes in an intersection and a local database where vehicles' intentions are collected. Then, while approaching the intersection, each vehicle queries the database to check if any of the surrounding vehicles will simultaneously be at the same virtual collision point. All the aforementioned approaches use the current vehicle information to estimate the future vehicles' pair-wise distances or collision times, assuming, however, only constant movement, without any speed or direction changes. As a consequence, these methods are reactive in nature and are inherently affected by a delay in detecting imminent collisions at urban intersections.

Very few approaches, e.g., \cite{Tao_TPCD, Wang_TPCD, marco_CA, Ruifeng_TPCD}, belong to the second class of collision detection algorithms. As in our work, in this class of solutions, more precise trajectory predictions are used as a mean to foresee or prevent collisions. In \cite{Ruifeng_TPCD, Wang_TPCD}, a KF is used to predict the future trajectories of the vehicle, with TTC used as a risk factor to identify the potential collisions. Similarly, \cite{Tao_TPCD} uses a non-linear Square-Root Unscented Kalman Filter (SR-UKF) to predict vehicles' trajectories, and utilizes the Monte Carlo sampling method to define the uncertainty region around the predicted trajectories. Then, a set of collision risk factors, such as the probability of the current trajectory, TTC, and conflict points in vehicles’ bounding boxes, are used to detect collisions.
 
A different approach is followed in \cite{CA_VD}, where a decentralized collision avoidance method is proposed. Specifically, this method forbids a vehicle to enter specific sets of speed and position configurations that eventually lead to collisions. Based on the assumption that the vehicle path and collision area are known, the study predicts if a vehicle is about to enter such a set, based on a Kalman filter prediction technique, and, if this is the case, it ensures that the automated collision avoidance mechanism located on-board the vehicle executes specific strategies to avoid the imminent danger. Similarly, \cite{KFTA} and \cite{CAMethods} account for constant turn rate and acceleration-based unscented Kalman filter to predict collisions. Accounting for human drivers, herein the automated collision avoidance mechanism enters in action only if the danger is imminent, and the driver has not intervened as of yet. Furthermore, \cite{marco_CA} approximates future vehicles' trajectories by solving a fourth-degree equation concerning the vehicles' derivatives (position, velocity, and acceleration). Collisions are then determined based on the time (time-to-collision) and spatial (space-to-collision) domain thresholds across all vehicle pairs.

Finally, following a completely different philosophy, in the state of the art there exists a set of approaches that, instead of predicting a collision, they plan safe trajectories for the vehicles based on an estimation of risk from their surroundings (\cite{pathplan,pom}). Nevertheless, this class of solutions only applies to autonomous vehicles, hence it is outside of the scope of our work.

\subsection{Novel Contribution}
\label{subsec:related_contribution}
Overall, even though there exists a large body of work concerning vehicle trajectory prediction and LSTM models have already been used for this task, e.g., \cite{altche2017lstm}, our framework differs from  prior art as it exploits a holistic view of the area of interest, including the past and the present status of both the infrastructure and all the vehicles crossing that area. Such holistic view is obtained thanks to the adoption of 5G and MEC technologies. Furthermore, differently from any other work in the field, prediction intervals are used to measure the reliability of the trajectory predictions. This allows capturing the presence of multiple possible future driver choices in a single metric, which can be computed efficiently, since it does not require modeling separately all maneuvers.

To the best of our knowledge, this work presents, for the first time, an ML-aided uncertainty-aware framework that allows obtaining efficient preemptive collision detection, even when considering human-driven vehicles at urban intersections. Indeed, contrary to all proposed approaches, our collision forecast method exploits both precise trajectory predictions and the associated uncertainty estimations, which allow raising alarms, in specific scenarios, even if the vehicles' predicted trajectories appear to be safe. Thanks to the use of the RFC, the adopted framework is able to easily recognize  real-time patterns of trajectory predictions and uncertainty estimations that led to dangerous situations in the past, raising the corresponding alarms within tight latency constraints. As a result, the proposed framework is able to predict an imminent danger well in advance, so that human drivers can avoid the collision by performing  simple  maneuvers.

Using real-world data, the proposed framework shows improved performance in all monitored metrics (i.e., in terms of trajectory prediction errors, collision detection false positives, available reaction time to drivers to avoid an imminent collision) when compared to: (i) the most relevant work on collision detection at urban intersections, i.e., \cite{marco_CA}; and (ii) a collision detection technique based only on trajectory predictions. In particular, when compared to a conventional technique that considers only point-based trajectory predictions, our framework improves the median available reaction time by $61\%$. Hence, given the obtained performance, our framework satisfies all the requirements (highly accurate output and short runtime) of the collision avoidance application at hand.

\begin{figure}[!ht]
    \centering
    \includegraphics[width=0.825\columnwidth]{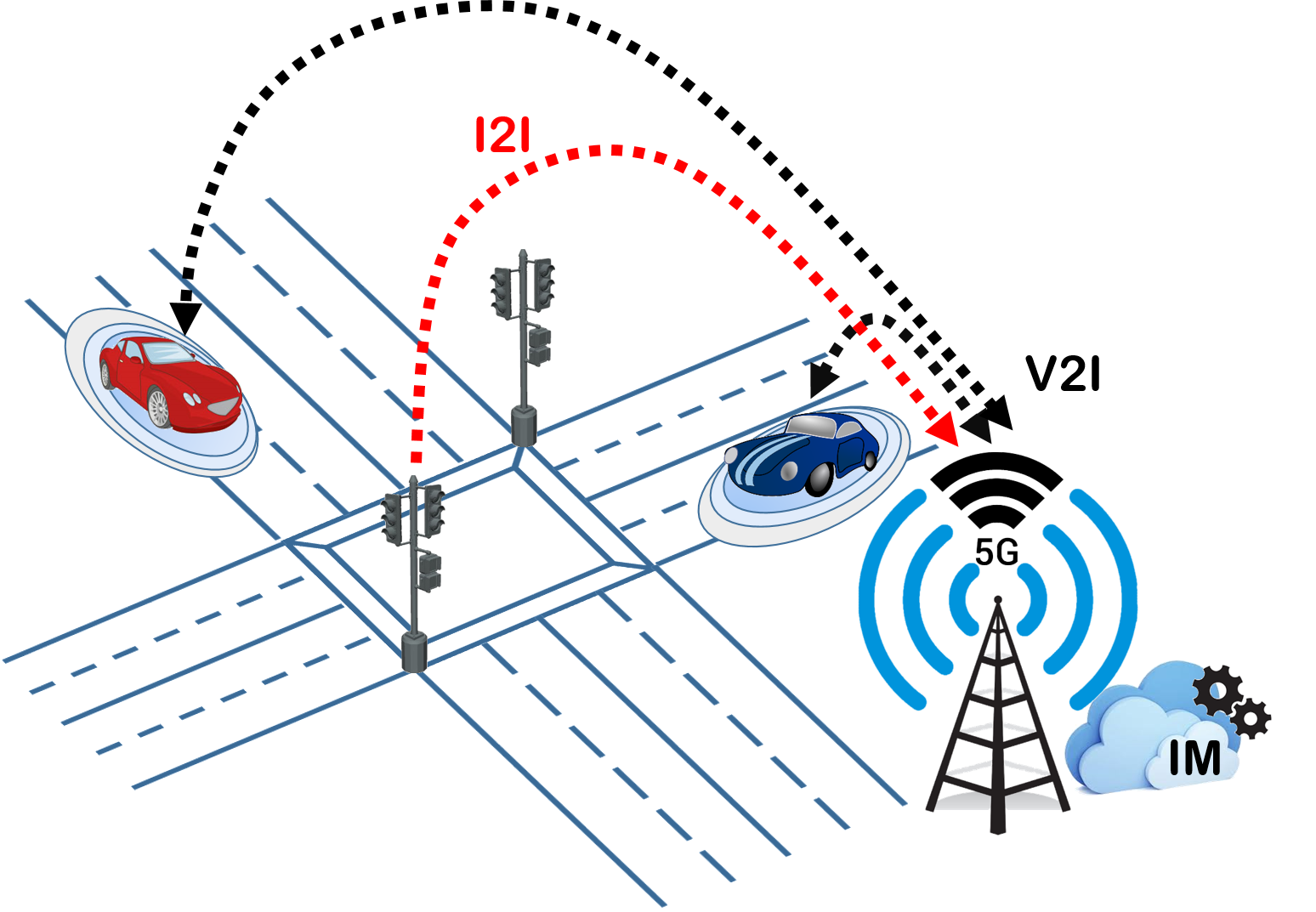}
     \caption{An edge computing-based scenario, including an IM collecting data generated by the vehicles and the road infrastructure and transmitting alarms in case of imminent collisions.}
    \label{fig:system_model}
    %\vspace{-6mm}
\end{figure}

\section{Network-Assisted Collision Detection: overview of our framework}
\label{sec:system_model}
In this work, we focus on a network-assisted preemptive collision detection framework that allows vehicles to prevent dangerous situations. Although the use of such an application can be extended to any dangerous area of the road infrastructure, here we focus on automated and human-driven CVs crossing urban intersections, as depicted in Fig.\,\ref{fig:system_model}. 

In our framework, an IM is hosted at the MEC \cite{avino2019mec} in the proximity of a gNodeB (gNB) covering the geographical area around the intersection. The appropriate number of computing resources to allocate to the IM to respect the time-sensitive nature of the application at hand can be determined through strategic scheduling even in a MEC platform running multiple services, as shown in \cite{Brik_RA}. Through integrated V2I and V2V communications, the IM can gather data from multiple sources, namely, (i) onboard sensor measurements sent by the vehicles crossing the intersection through Cooperative Awareness Messages (CAMs), and (ii) infrastructure-based information transmitted by city smart sensors and cameras. Note that CAMs can include several pieces of information provided by the vehicle's Controller Area Network (CAN) bus, including speed, direction, steering angle, acceleration, braking, yaw rate, and relative distances with surrounding vehicles obtained through LiDAR. As for the data collected by the road infrastructure, these can include traffic light phases, vehicle lane information, and number of vehicles in each lane. As a result, the benefits of using a central entity, i.e., the IM, at the network edge for road safety are multi-fold: (i) at any time, the IM has a significantly richer view of the intersection than the individual vehicles, (ii) the IM can collect CAMs and historical data (e.g., collisions/dangerous situations) in a local database to train precise models, and (iii) unlike cloud implementations, the IM meets the tight latency constraints of real-time safety applications. In this work, with the obtained extended view of the intersection, the IM accurately predicts potential future dangerous situations. Moreover, upon detecting two or more vehicles on a collision course, it promptly alerts them by transmitting a Decentralized Environmental Notification Message (DENM). Upon receiving such an alarm, CVs will start braking; thus, the objective of the IM is to detect a possible upcoming collision early enough to let vehicles avoid the imminent risk.

\begin{figure*}
    \centering
    \includegraphics[width=1.0\textwidth]{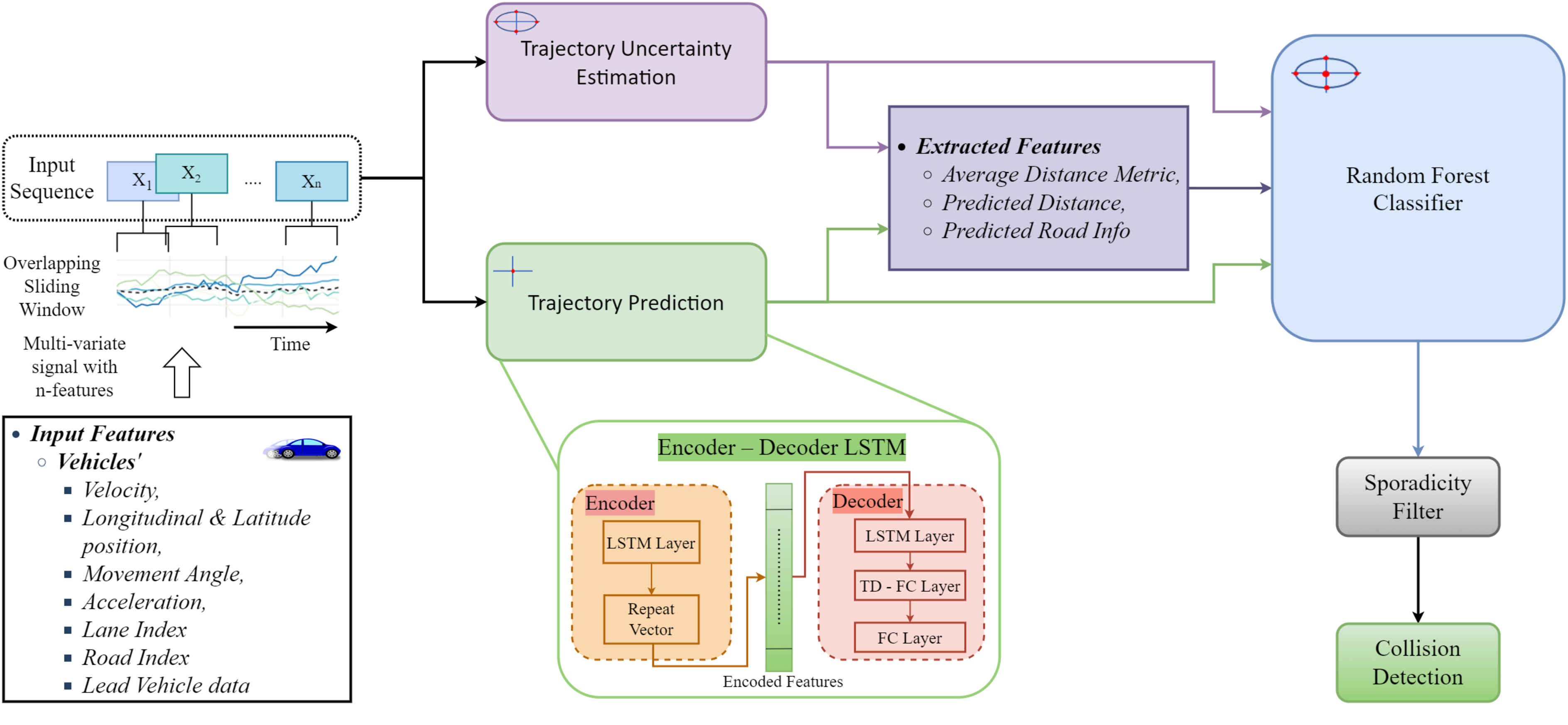}
     \caption{An overview of the proposed network-assisted collision detection methodology.}
    \label{fig:framework}
    %\vspace{-6mm}
\end{figure*}

To this end, we envision the implementation at the IM of the collision detection framework whose structure is depicted in Fig.\,\ref{fig:framework}. Our detection framework searches for a possible collision as soon as a new CAM, updating the information relative to a vehicle status, is received. Using a sliding window approach, recent past information originating from vehicles and the road infrastructure are used as \textit{input data} to our framework. Collision detection at intersections is in essence composed of two complementary, yet distinct, sub-problems that operate in cascade, i.e., the latter takes as input the output of the former: (i) trajectory prediction and (ii) detection of vehicles on a collision course. \textit{Vehicle trajectory prediction} consists of point-based estimates, inevitably containing some uncertainty. Thus, to develop a trustworthy collision detection mechanism, \textit{trajectory prediction uncertainty} is appropriately measured, interpreted, and processed before decision-making. For both trajectory prediction and uncertainty estimation (even though with different loss functions), LSTM-ED models are exploited to: (i) solve the problem of storing long time-steps in the learning memory~\cite{279181}; (ii) capture the intrinsic non-linear nature of vehicle trajectories; and (iii) scale to high-dimensional data (unlike Bayesian learning).

The obtained vehicles' trajectory predictions, along with the associated uncertainty estimation and other relevant information, are subsequently used by the IM to train an RFC that, based on historical data, recognizes patterns of dangerous inputs dangerous inputs so to detect in real-time the event of a collision or of situations that are deemed dangerous. Finally, to avoid spurious detections, the IM employs a \textit{sporadicity filter} that transmits to the vehicles involved an actual alarm only if, for a specific pair of vehicles, the RFC detects a dangerous situation for a number of consecutive predictions. Overall, a novel uncertainty-aware collision detection mechanism is developed, showing, as presented in Sec. \ref{sec:peva}, to be capable of accurately and timely identifying  future collisions. Interestingly, adding to the proposed framework a measure of the trajectories' certainty level is crucial for improving the obtained performance.

\section{Vehicle Trajectory Prediction and Uncertainty in Deep LSTM-ED}
\label{sec:LSTM_network}

The trajectory prediction sub-problem can be in general described as a sequence-to-sequence problem where the past vehicle trajectory is used as input, together with other relevant information, for predicting the future vehicle trajectory for several time-steps ahead. LSTM-ED models are a promising modeling approach (as shown in~\cite{Dinesh21}) for the trajectory prediction problem. As in similar use cases, a least squares loss function is minimized to find a predictive model that, in essence, approximates the relationship between the input data collected by the IM and the future vehicle positions. Hence, the output of the model is an estimate of the future vehicles' positions, inevitably containing some uncertainty. This estimate is in fact the mean response value, with the variability around that mean representing the uncertainty of the model itself and of the noisy input data.

When collision avoidance is the ultimate goal of the trajectory prediction sub-problem, a point-estimate is not enough, as it may lead to erroneous collision detections or more severely to the underestimation of a possible collision event. While the latter situation is more severe as the vehicles cannot be notified timely towards collision avoidance, the former should not be ignored, since it may lead to inappropriate decisions/actions with many undesired effects (i.e., frustration of drivers or even collisions). Measuring the certainty level of trajectory estimates, essentially building prediction intervals that quantify the existence of multimodal future trajectories~\cite{Meeker17}, helps reducing the errors associated with our framework since: (i) it reduces the number of false collision detections when the average trajectory estimations suggest a collision, but a large prediction uncertainty is associated with them; and (ii) it avoids underestimating a possible threat when a large uncertainty is associated with safe trajectory predictions.

To this end, in this section we focus on trajectory prediction (Sec.\,\ref{sec:trajectory_estimation}) and uncertainty estimation (Sec.\,\ref{sec:uncertainty_estimation}) with the purpose of defining a set of appropriate, uncertainty-aware trajectory estimates, subsequently utilized by the collision detection mechanism developed in this work. Furthermore, the set of inputs used by the presented model is introduced in Sec.\,\ref{dataset_formulation}. The LSTM-ED model used for both trajectory prediction and uncertainty estimation is then described in Sec.~\ref{lstm_model}.

\subsection{Least Squares Regression for Trajectory Prediction: Estimating Mean Response Values}
\label{sec:trajectory_estimation}
For trajectory prediction, a least squares LSTM-ED model is trained to minimize the Mean Squared Error (MSE) loss function given by:
\begin{equation}
L_{mse}=\frac{1}{NL} \sum_{i=1}^N  \sum_{j=1}^{L} || {\bf y}_{i,t+j} -  \hat{\bf y}_{i,t+j} ||^2_2,
\label{mse_loss} 
\end{equation} 
where $N$ is the number of vehicle sequences used for MSE evaluation (i.e., during training/testing), $L$ are the time-steps used for prediction, $t$ is the present time, ${\bf x}_i \in \mathbb{R}^{d}$ is a vector describing vehicle $i$'s input data, ${\bf y}_{i,t+j} \in \mathbb{R}^2$ is a vector describing the true location (i.e., latitude and longitude information) of vehicle $i$ at time slot $t+j$, 
and $\hat{\bf y}_{i,t+j} \in \mathbb{R}^2$ is the estimated vector inferred by the trained least squares LSTM-ED model $\hat{F}$. In essence, $\hat{F}$ is formally defined as ${\hat F}({\bf x}_i, \{\hat{\bf y}_{i,t+s}\}_{s=0}^{j-1})$ as it utilizes as inputs both present and past data, i.e., ${\bf x}_i$, and previously predicted trajectory values to infer the trajectory value for the next step ahead. Hence, when $j=1$, then ${\hat F}({\bf x}_i)$ predicts $\hat{\bf y}_{i,t+1}$, with this estimate subsequently used as input to $\hat{F}({\bf x}_i, \hat{\bf y}_{i,t+1})$ to predict $\hat{\bf y}_{i,t+2}$, and so on.

\subsection{Deep Quantile Regression for Trajectory Prediction: Estimating Lower and Upper Response Values}
\label{sec:uncertainty_estimation}
As previously mentioned, our uncertainty estimation method is based on building prediction intervals.
By definition, a prediction interval is a range of values in which a single future observation will fall in, with a certain probability, based on the estimated model (i.e., given the previous observations used for training the model). While in ML several approaches exist for building prediction intervals, such as conducting Bayesian or Monte Carlo (MC) dropout inference~\cite{Gal16}, in this work, a deep quantile regression framework is adopted~\cite{Rodrigues2020}, providing a principled way of controlling the certainty level of the estimates. Specifically, a deep LSTM-ED model is trained to estimate a set of conditional quantile functions capable of providing lower and upper estimates for each future vehicle position. 

To this end, we apply deep quantile regression to approximate the conditional quantile function $y^q_{i,t+j} = Q_q ({\bf x}_i,\{{y}^q_{i,t+s}\}_{s=1}^{j-1})$, also known as the $q$-quantile, where $q$ is the probability that the random variable $y_{i,t+j} \in {\bf y}_{i,t+j}$ (representing the true vehicle location at the $t\mathord{+}j$-th future time step) conditional on the value of the random variable ${\bf x}_i$ (representing the input data) is less than or equal to $y^q_{i,t+j}$, i.e.,  $\PP(y_{i,t+j} \leq y^q_{i,t+j}|{\bf x}_i,\{{y}^q_{i,t+s}\}_{s=1}^{j-1})=q$ \cite{Koenker05}. 

To derive lower and upper vehicle trajectory estimates, two $q$-quantiles need to be estimated. Since the objective is to create a prediction interval able to contain most of the true future vehicle locations given the input data ${\bf x}_i$, a low enough quantile (i.e., with $q \rightarrow 0$) and a high enough quantile (i.e., with $q \rightarrow 1$) must be used for the lower and upper quantiles, respectively (denoted as $l$ and $u$, respectively). It is worth mentioning, that while for each $q$ value different quantile models can be trained, in this work we opted to train one single model for both lower and upper estimates of the vehicle's future position (i.e., longitude and latitude). Hence, an LSTM-ED model is trained to return upper and lower estimates for the longitude future locations of the vehicle, and an LSTM-ED model is trained to return upper and lower estimates for the latitude future locations of the vehicle. Considering that, in a deep learning framework a $q$-quantile is estimated by minimizing the asymmetrically weighted sum of absolute errors~\cite{Rodrigues2020, Koenker05}, the loss function minimized during training is given by:

\begin{equation}
L_q = \frac{1}{2NL} \sum_{i=1}^N  \sum_{j=1}^{L}  {\rho}_{l}{\big(} y_{i,t+j} - \hat{y}^l_{i,t+j} {\big)}+{\rho}_{u}{\big(} y_{i,t+j} - \hat{y}^u_{i,t+j} {\big)}, 
\label{eq_quantile_loss} 
\end{equation}
where $q$ denotes any of the two quantiles $u$ and $l$:
\begin{equation}
\rho_{q}(z) = \Bigg\{
        \begin{array}{ll}
         q z, & \text{if } z \geq 0,\\
        (q -1) z, & \text{if } z < 0, 
        \end{array}
\end{equation}
and where $\hat{y}^q_{i,t+s} \in \hat{\bf y}^q_{i,t+s}$ is the estimation of the trained $q$-quantile LSTM-ED model $\hat{Q}_q$. Similarly to $\hat{F}$, in this work the approximation of $Q_q$ obtained with the LSTM-ED model, i.e., $\hat{Q_q}$, sequentially utilizes as inputs previous quantile values $\{\hat{y}^q_{i,t+s}\}_{s=1}^{j-1}$. Hence, when $j=1$, then ${\hat Q}_{q}({\bf x}_i)$ returns $\hat{y}^q_{i,t+1}$, subsequently used as input to $\hat{Q}_q (x_i, \hat{y}^q_{i,t+1})$ to return $\hat{y}^q_{i,t+2}$, and so on. 

\subsection{Dataset Formulation}~\label{dataset_formulation}
For both the considered least squares and quantile loss functions, the objective is to optimize the unknown parameters of an LSTM-ED model such that a non-linear function (i.e., the $F$ or $Q_q$ function, respectively) is approximated. In the former case, the approximation allows accurately predicting the vehicle's future location:
\begin{equation}
{\bf y}_i=[{\bf y}_{i,t+1}, {\bf y}_{i,t+2}, \cdots, {\bf y}_{i,t+L}]\,
\end{equation}   
where
\begin{equation}
{\bf y}_{i,t+j}=[y_{i,t+j}^{(\lambda)}, y_{i,t+j}^{(\phi)}] 
\end{equation}
is a vector including latitude vehicle information $y_{i,t+j}^{(\lambda)}$ and longitude vehicle information $y_{i,t+j}^{(\phi)}$ at the $j$-th future time step. In the latter case, instead, the approximation allows to accurately estimate the prediction intervals of the vehicle's future location:
\begin{equation}
{\bf y}^q_i=[{\bf y}^q_{i,t+1}, {\bf y}^q_{i,t+2}, \cdots, {\bf y}^q_{i,t+L}]\,
\end{equation}   
where
\begin{equation}
{\bf y}^q_{i,t+j}=[y_{i,t+j}^{l}, y_{i,t+j}^{u}] 
\end{equation}
is a vector that includes the lower and upper estimates of the latitude/longitude vehicle location at the $j$-th future time step (i.e., $y_{i,t+j}^{l}$ and $y_{i,t+j}^{u}$, respectively).

To make predictions, the model is given as input the vehicle's past and present information:
\begin{equation}
{\bf x}_i=[{\bf x}_{i,t-T}, {\bf x}_{i,t-T+1}, \cdots, {\bf x}_{i,t-1}, {\bf x}_{i,t}].
\end{equation}
In the aforementioned expression, $T$ is the number of past and present observations, $t$ is the present time instant, $L$ is the number of next vehicle observations (i.e., the prediction horizon), and ${\bf x}_{i}$, ${\bf y}_i$, and ${\bf y}^q_i$ form a time series (i.e., are sequential in time), with each time instant of interest being $\tau$ seconds apart (i.e., $\tau$ is the time scale of sampling/estimating time instants). Regarding the past vehicle input data ${\bf x}_i$, each ${\bf x}_{i,j}$, $\forall j=t-T,...,t$, is a vector containing $d$ measurements for both the on-board and infrastructure components. Specifically, each ${\bf x}_{i,j} \in \mathbb{R}^d$ consists of the following set of observations:
\begin{equation}
{\bf x}_{i,j}  \mathord{=} \{x^{(\lambda)}_{i,j}\!, x^{(\phi)}_{i,j} \!, x^{(r)}_{i,j}\!,  x^{(v)}_{i,j}\!, x^{(a)}_{i,j}\!, x^{(l)}_{i,j}, x^{(e)}_{i,j},  x^{(p,\lambda)}_{i,j}\!\!\!, x^{(p,\phi)}_{i,j} \!\!\!, x^{(p,v)}_{i,j}\},
\end{equation}
where:
\begin{itemize}
\item $x^{(\lambda)}_{i,j}$ is the latitudinal position of vehicle $i$ at time instant $t-j$;
\item $x^{(\phi)}_{i,j}$ is the longitudinal position of vehicle $i$ at time instant $t-j$;
\item $x^{(r)}_{i,j}$ is the movement angle of vehicle $i$ at time instant $t-j$;
\item $x^{(v)}_{i,j}$ is the velocity of vehicle $i$ at time instant $t-j$;
\item $x^{(a)}_{i,j}$ is the acceleration of vehicle $i$ at time instant $t-j$;
\item $x^{(l)}_{i,j}$ is the lane index where vehicle $i$ is traveling on at time instant $t-j$;
\item $x^{(e)}_{i,j}$ is the road index on which vehicle $i$ is traveling at time instant $t-j$;
\item $x^{(p,\lambda)}_{i,j}$ is the latitudinal position of the vehicle preceding vehicle $i$ in the same lane, at time instant $t-j$;
\item $x^{(p,\phi)}_{i,j}$ is the longitudinal position of the vehicle preceding vehicle $i$ in the same lane, at time instant $t-j$;
\item $x^{(p,v)}_{i,j}$ is the velocity of the vehicle preceding vehicle $i$ in the same lane, at time instant $t-j$. 
\end{itemize} 

Amongst the input features, the lane index information ($x^{(l)}_{i,j}$) and road index information ($x^{(e)}_{i,j}$) are converted into a one-hot encoded vector to benefit the learning process. Rather than giving semantic information on the road structure, in our work, lane and road index help the LSTM-based trajectory prediction to identify the intention of drivers of turning or of proceeding straight at the intersection. Indeed, given the structure of an intersection, drivers can turn only when approaching the intersection on specific lanes of the road. Furthermore, apart from features describing the past movement of vehicle $i$, e.g., its location, speed, and direction, a set of information regarding the preceding vehicle on the same lane is also included. The rationale is that, at an intersection, the correlation between vehicle $i$'s future locations and the present location of the preceding vehicle is very high. Finally, it is worth mentioning that additional input features were also considered and tested during model training and inference, without however contributing to further improving  model accuracy. Even though our model achieves a very high accuracy, as shown in the following sections, a further analysis of input features naturally present in CAM messages (or at the infrastructure) constitutes an interesting future research direction.

\subsection{The LSTM-ED Model}\label{lstm_model}
\begin{figure}[h!]
\begin{center}
\includegraphics[scale=0.33]{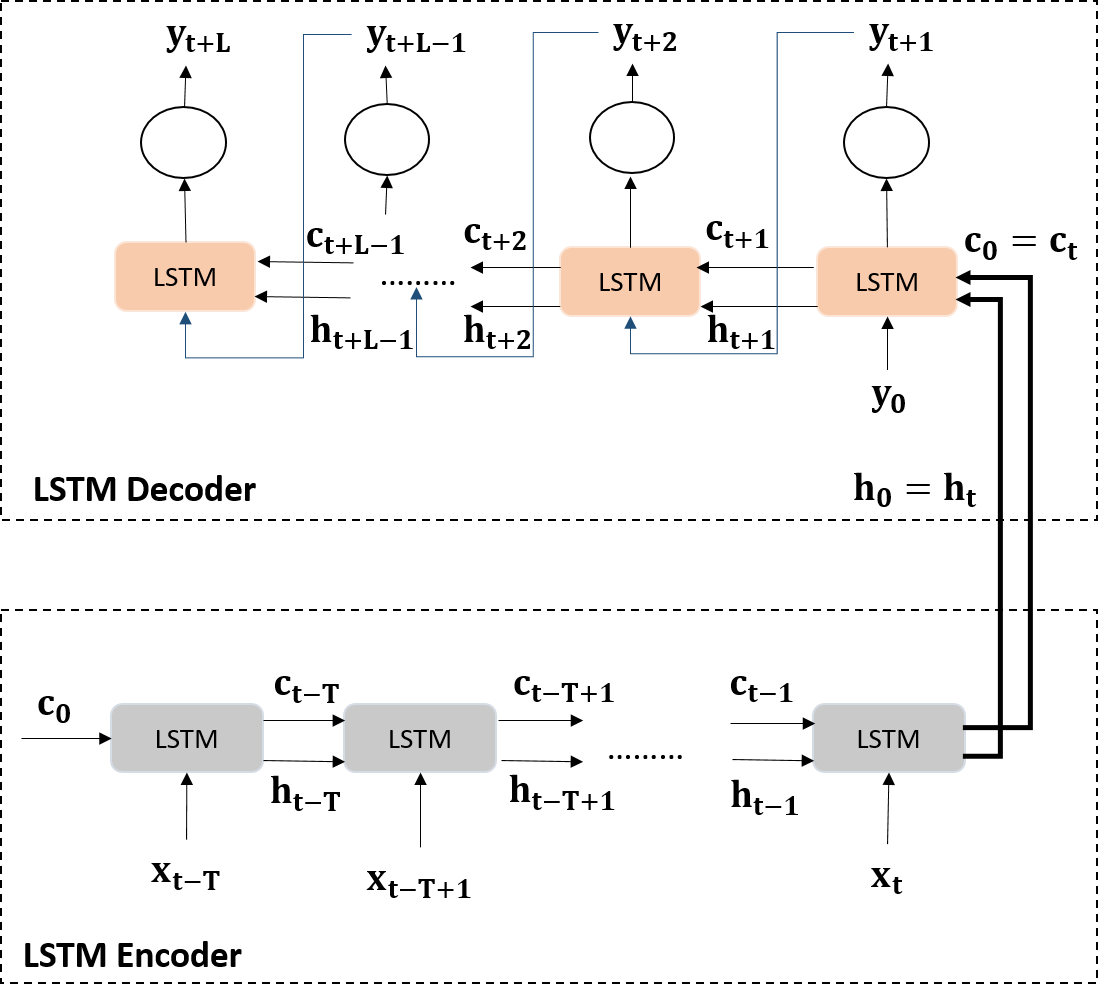}
\vspace{-1.5mm}
\caption{The encoder-decoder LSTM architecture used for vehicle trajectory estimation and uncertainty estimation.}
\vspace{-4mm}
\label{enc_dec}
\end{center}
%\vspace{-6mm}
\end{figure}

The LSTM-ED architecture we use is illustrated in Fig.~\ref{enc_dec}, where index $i$ is omitted to simplify the notation and the output ${\bf y}_{t}$ is alternatively equal to the trajectory prediction or the uncertainty estimation depending on the use of the LSTM model. Note that, for simplicity, encoders and decoders of one layer only are depicted, but the model can include more hidden layers (i.e., stack of LSTM layers). Both encoder and decoder components are formed by LSTM cells, which in Fig.~\ref{enc_dec} appear unfolded through time to depict their operation over the input and output sequences. In reality, inputs are processed sequentially, and outputs are obtained one after the other (and reintroduced at the input of the decoder for predicting the next time step). 

The key of LSTM is the cell state ${\bf c}_{t}$, that acts as the cell memory, storing a summary of the past input sequence. At the output, the LSTM cell combines such accumulated memory with the input received, so as to obtain the hidden state ${\bf h}_{t}$, also representing the output vector of the LSTM cell. At the decoder cell, the output ${\bf h}_{t}$ is fed to a fully connected dense layer (applied sequentially) to generate the output of the LSTM-ED model. Further details on the LSTM cell structure can be found in \cite{doi:10.1162/neco.1997.9.8.1735,Dinesh21}.

To summarize, the LSTM-ED model, parameterized by the set of unknown parameters of the model, learns a nonlinear function that predicts the current output value ${\bf y_{t+k}}$ as:
\begin{equation}
{\bf y}_{t+k}= \hat{\Theta}({\bf x}_{t-T}, \cdots, {\bf x}_{t-1},{\bf x}_t, {\bf y}_{t+1}, {\bf y}_{t+2}, \cdots, {\bf y}_{t+k-1}),
\end{equation}  
with the encoder providing the summary of the input data $\{{\bf x}_{t-k'}\}_{{k'}={0}}^T$ through the cell and hidden states ${\bf c}_t$ and ${\bf h}_t$, respectively, and $\hat{\Theta}$ being $\hat{F}$ or $\hat{Q}$, depending on the regression task at hand.

The two components of the LSTM-ED are trained on a dataset $D=\{{\bf x}_i,{\bf y}_i\}_{i=1}^N$ where $N$ is the number of labeled vehicle sequences, to minimize the MSE or quantile loss function. Note that, for what concerns the prediction intervals, the LSTM-ED model is trained for obtaining the uncertainty estimation on both latitude and longitude. Hence, dataset $D$ is further partitioned into two datasets $D^{(\lambda)}=\{{\bf x}_i,y^{(\lambda)}_i\}_{i=1}^N$ and $D^{(\phi)}=\{{\bf x}_i,y^{(\phi)}_i\}_{i=1}^N$. For training, the Adam optimization algorithm~\cite{kingma2017adam} is used. 
The actual implementation of the presented LSTM-ED model is available online and can be found at \url{https://github.com/krish-din/C-AVOID}.

\section{An ML-aided uncertainty-aware collision detection mechanism}\label{sec:decision_tree}
Because of our LSTM-ED approach, for any vehicle in the area of interest, both the trajectory prediction and the associated uncertainty estimation are available at the IM for a prediction window of $L$ time-steps. The IM exploits such estimation of the future vehicles' position, along with other relevant data, to make an informed decision on whether a pair of vehicles is on a collision course. To this end, the IM trains an RFC \cite{Breiman2001,louppe2015understanding} with the aim to determine if two vehicles, described by an input pattern $\chi = \{ \chi_1, \chi_2, ..., \chi_k\}$, are expected to collide within the prediction window, in which case the RFC returns classification outcome $\upsilon=1$, otherwise it returns $\upsilon=0$.

In a nutshell, RFC is an ensemble of multiple independent decision trees, where the classification of unseen patterns is performed through majority voting.  Thanks to its structure, RFC is one of the most successful ensemble learning techniques,robust to  
overfitting (as compared to individual decision trees \cite{Breiman2001}), and proven to be very powerful 
for problems characterized by high dimensionality and  by skewed data. Thus, it suits very well 
our scenario where real-world data is expected to be skewed, due to the presence of outlying 
driving behaviors. 

Specifically, we apply feature subsampling  to build the trees by splitting the training dataset 
into sub-datasets according to the value taken by the input features. Eventually, each decision tree 
is organized into branches composed of two types
of nodes, i.e., the decision nodes (at which splits are made) and the leaf nodes, where the 
classification outcome  takes place. Finding the best split at the decision nodes involves choosing 
the feature and the split value (i.e., a threshold)
for that feature that will result into the highest improvement to the model, given the desired outcome $\upsilon$. 
That is, at each split, two sub-datasets with lower “impurity” are obtained
when compared to the original dataset at the decision node. In this work, similarity within each 
sub-dataset is computed following the Gini Index (GI) \cite{louppe2015understanding},
which measures the portion
of samples in each sub-dataset having different outcomes. At each decision node, since the number 
of split options is large, only a subset of potential split values is considered per feature 
\cite{ho1998random}. 

Splitting at the decision nodes continues until either all values in the sub-datasets of 
that node are pure (i.e., all samples describe a collision or a non-collision) or some other
conditions are met (e.g., maximum tree depth).

In the inference phase, to classify any unseen pattern, the tree is navigated up to reaching a 
leaf node containing an output for the objective binary variable $\upsilon$. A specific leaf node is 
reached if the unseen pattern falls into specific ranges for the input features, as learned 
during training.

\subsection{RFC's Inputs}~\label{dataset_formulation_RF}
As it is clear from the random forest approach presented above, the classification of the input data relative to two vehicles $i$ and $k$ in the ``collision'' class happens if the chosen input is within specific features' intervals. Hence, the choice of the input features is critical for improving the obtained performance. 

Among the different possible input features, present and future vehicle locations certainly play a crucial role in determining whether the two vehicles are on a collision course.  Indeed, such features help to determine specific locations on the road infrastructure where more often vehicles tend to collide/experience dangerous situations. Accordingly, also the road direction traveled by the vehicles under analysis carries important information, since collisions are usually prone to happen to vehicles heading towards a specific route. 

Intuitively, the future predicted distance between vehicles is also an indicator of a possible imminent collision. Indeed, independently from their location, a dangerous situation is certain in case two vehicles are predicted to be too close to each other. From the experiments performed, two metrics of proximity between vehicles helped to obtain the best performance, one based on the trajectory prediction and one based on its uncertainty estimation. That is, the first one is the Euclidean Distance (ED) $d_{ik,t+j}$ between the predicted location of vehicles $i$ and $k$ at time $t+j$. The second one is based on transforming the obtained prediction intervals into a distribution around the trajectory prediction described in Sec.\,\ref{sec:trajectory_estimation}, and on computing in distribution (and not based on the point estimates) the squared expected distance $E[d_{ik,t+j}^2]$ between the predicted location of vehicles $i$ and $k$ at time $t+j$. Specifically, it is assumed that the uncertainty associated with the trajectory prediction follows a Gaussian distribution, an assumption that is typically valid when the uncertainty is due to a combination of several independent factors, as in the use case under consideration. Then, the future location of vehicle $i$ follows the multivariate Gaussian distribution $\mathcal{N}({\bf y}_{i,t+j},\Sigma_{i,t+j}=[\sigma_{i,t+j}^{2~~(\lambda)}~~0; 0~~\sigma_{i,t+j}^{2~~(\phi)}])$. Given that the length of the prediction intervals is known from the method introduced in Sec.\,\ref{sec:uncertainty_estimation}, the covariance matrix of the Gaussian distribution can be easily obtained as follows\cite{ribeiro2004gaussian}:
\begin{equation}
\sigma_{i,t+j}^{2~~(\lambda)}=\frac{(y_{i,t+j}^{u(\lambda)}-y_{i,t+j}^{l(\lambda)})^2}{K_\epsilon};~~~ \sigma_{i,t+j}^{2~~(\phi)}=\frac{{(y_{i,t+j}^{u(\phi)}-y_{i,t+j}^{l(\phi)})^2}}{K_\epsilon}
\end{equation}
where $\sigma_{i,t+j}^{2~~(\lambda)}$ and $\sigma_{i,t+j}^{2~~(\phi)}$ are the variances associated with the predicted latitude and longitude location, respectively, of vehicle $i$ at time $t+j$, and $K_\epsilon$ is the inverse of the Cumulative Density Function (CDF) of the chi-squared distribution having two degrees of freedom computed at $1\mathord{-}\epsilon=1-(u-l)$. As a result, the expected squared ED between the two vehicles $i$ and $k$ can be computed as:
\begin{equation}
E[d_{ik,t+j}^2] = ||{\bf y}_{i,t+j}-{\bf y}_{k,t+j}||^2 + \Trace[\Sigma_{i,t+j}+\Sigma_{k,t+j}]
\end{equation}
where  $\Trace[\cdot]$ denotes the trace operator. 

Finally, we also include in the input data the size of the prediction intervals and the covariance matrices $\Sigma_{i,t+j}$ and $\Sigma_{k,t+j}$ as a measure of the uncertainty estimation in order to account in the classification task for the reliability level of the information based on the trajectory predictions, i.e., the future positions of vehicles $i$ and $k$ and their ED. 

To summarize, the input data $\chi$ concerning a pair of vehicles $i$ and $k$ and a specific future time instant $t+j$, includes the following features:
\begin{itemize}
\item $y^{(\lambda)}_{z,t+j}$ ($y^{(\phi)}_{z,t+j}$) is the latitudinal (longitudinal) position of either vehicle $i$ or $k$ at future time instant $t+j$;
\item $x^{(l)}_{z,t+j}$ is the road direction where either vehicle $i$ or $k$ are predicted to be traveling on at the future time instant $t+j$;
\item $d_{ik,t+j}$ is the ED between vehicles $i$ and $k$ at future time instant $t+j$, based on the predicted point estimates;
\item $E[d_{ik,t+j}^2]$ is the expected squared ED between vehicles $i$ and $k$ at future time instant $t+j$, based on the predicted distribution estimates;
\item $y_{z,t+j}^{u(\lambda)}-y_{z,t+j}^{l(\lambda)}$ ($y_{z,t+j}^{u(\phi)}-y_{z,t+j}^{l(\phi)}$) is the latitudinal (longitudinal) prediction interval of either vehicle $i$ or $k$ at future time instant $t+j$;
\item $\Sigma_{z,t+j}$ is the covariance matrix associated with the predicted location ${\bf y}_{z,t+j}$ of either vehicle $i$ or $k$ at future time instant $t+j$.
\end{itemize} 

\section{Avoiding Collisions at Urban Intersections}
\label{sec:collision_avoidance}
Due to the proposed approach, the IM can exploit an uncertainty-aware collision detection mechanism to predict in advance whether two vehicles are on a collision course. Once a collision is detected, the IM informs the vehicles in danger through unicast DENMs, so that corrective actions can be undertaken. In order to reduce the number of alarms relative to situations that do not lead to an actual collision or to an actual dangerous circumstance (i.e., to reduce the number of unnecessary and sudden corrective actions), the IM transmits DENMs only after the RFC has detected a collision over a number of consecutive observations. 

\begin{figure}[h!]
\begin{center}
\includegraphics[width=0.825\columnwidth]{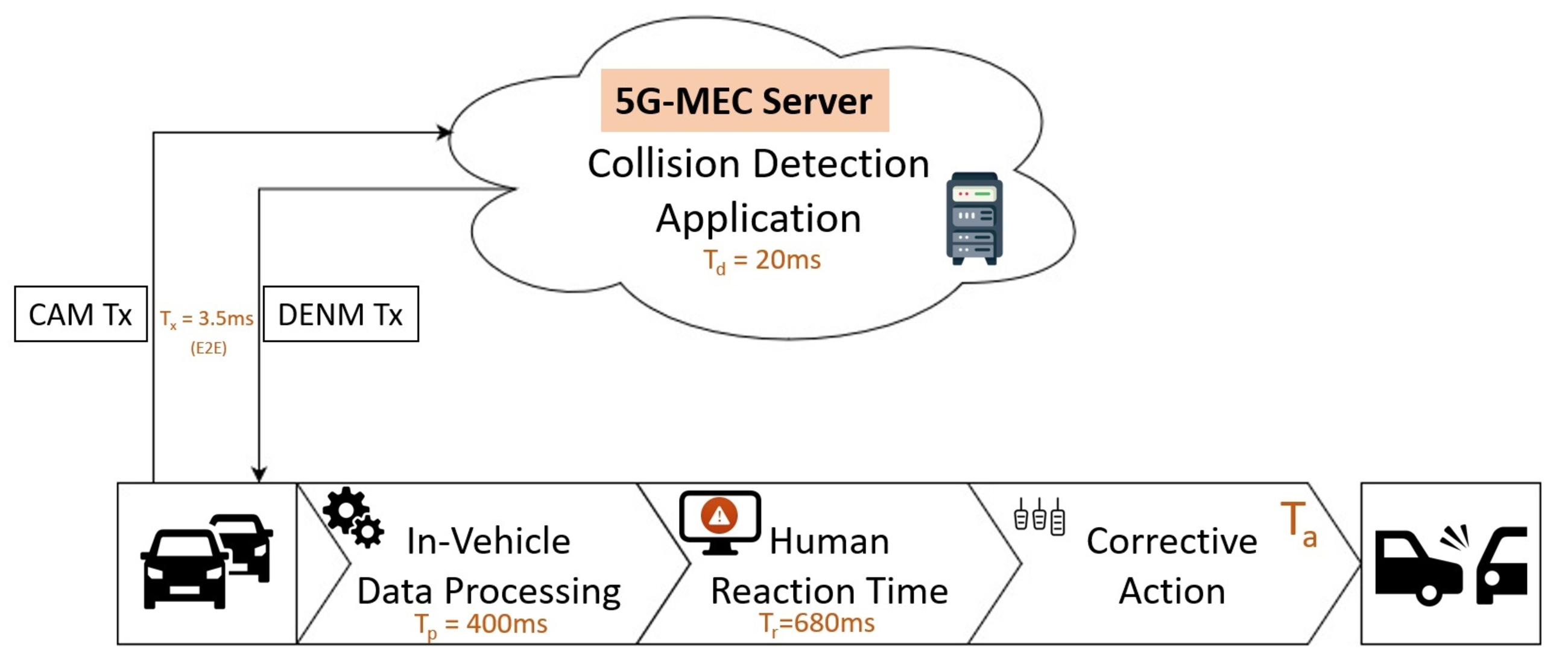}
\vspace{-1.5mm}
\caption{Collision avoidance - An overview of the latency components involved.}
\label{fig:latency}
\end{center}
\end{figure}

Selecting the right number of consecutive observations, which leads to a DENM message transmission, represents a crucial trade-off between detecting  a collision on time, so that corrective actions are effective, and reducing the number of generated false alarms. In this context, it has to be considered that, from the moment the DENM is transmitted to the moment a vehicle starts braking, several latency components need to be accounted for. Figure~\ref{fig:latency} illustrates an overview of such latency components. As an alarm is raised by a CAM  that triggers the detection of a collision at the IM, the end-to-end communication latency component, i.e., $T_x$, includes both CAM and DENM transmission latencies. To reduce this latency, as mentioned earlier, the IM runs at the edge on the 5G-MEC platform, geographically close to the intersection where the collision detection application is intended to work. The detection latency component, i.e., $T_d$, models the time required by the IM to process the information received from the vehicles. Thereafter, the DENM is received at the vehicle side and before a corrective action is actually undertaken, two further latency components need to be considered,  i.e., the in-vehicle processing time $T_p$ and the human reaction time $T_r$. Clearly, these latency components are modeled differently if a vehicle is assumed to be exclusively human-driven or if it is assisted by an automated braking system. In the former case, the two components include the fact that the vehicle processes the DENM and displays on the dashboard a visual alarm to the driver, which reacts to it with some delay. In the latter case, instead, the vehicle processes the DENM and automatically starts braking, without additional delay, i.e., $T_r=0$. As a corrective action, we assume that vehicles brake as soon as the alarm is correctly analyzed, with the intention of reaching a halt position (reached after braking for $T_a$ seconds) before colliding with any other vehicle in the intersection. Hence, in our system, a collision is considered as successfully detected if:
\begin{equation}
T_x+T_d+T_p+T_r+T_a<T_c,
\end{equation}
where $T_c$ is the time interval between the moment a CAM is transmitted by a vehicle and the moment such vehicle is involved in a collision. It must be noted that the different latency components represent random variables, which depend on the specific scenario at hand and which specific distribution is detailed in Sec. \ref{subsec:collision_avoidance} below.

\section{Performance Evaluation\label{sec:peva}}
In this section, we evaluate the proposed approach using realistic simulated mobility data traces for the cities of Luxembourg and Monaco.

\subsection{Data Collection and Pre-processing}
\begin{figure*}
\center
\includegraphics[width=70mm,scale=0.5]{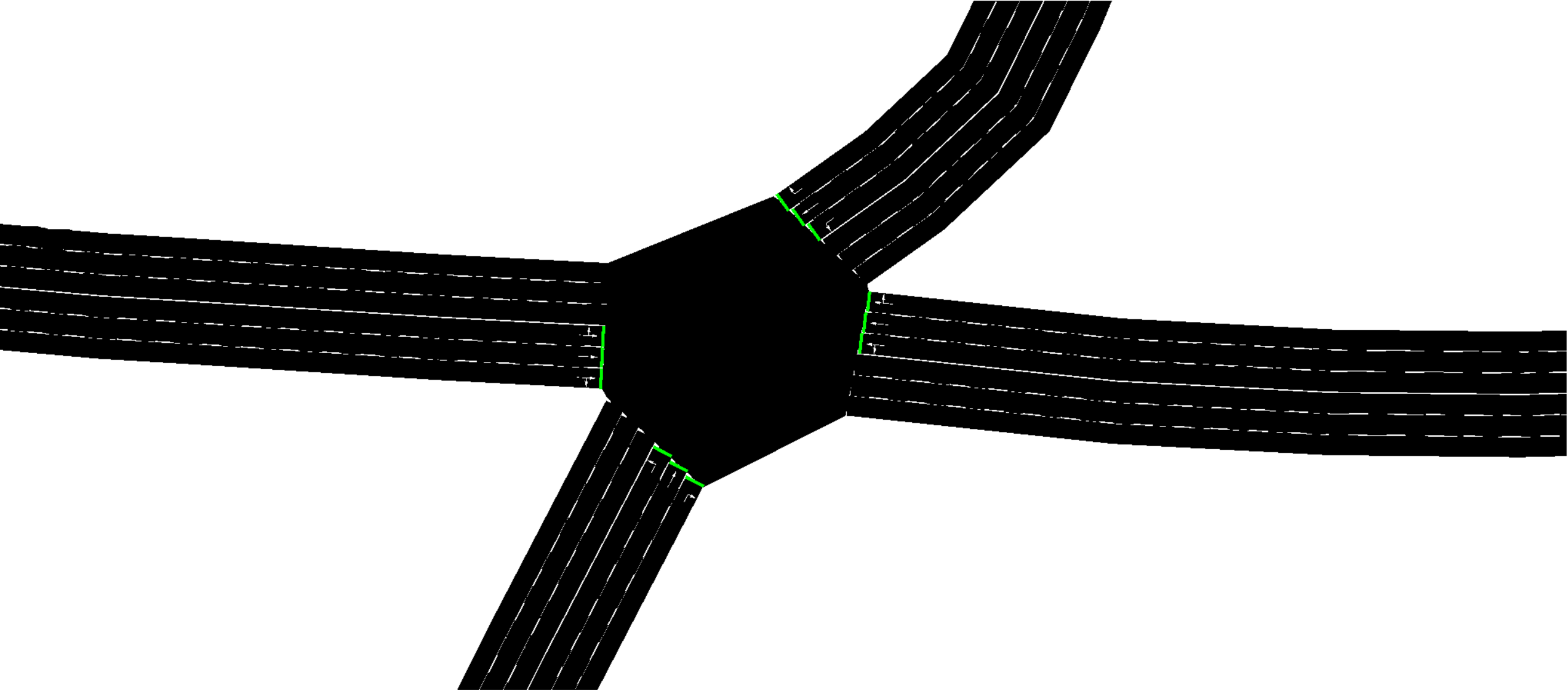}
\hspace{0.5cm}
\includegraphics[width=70mm,scale=0.5]{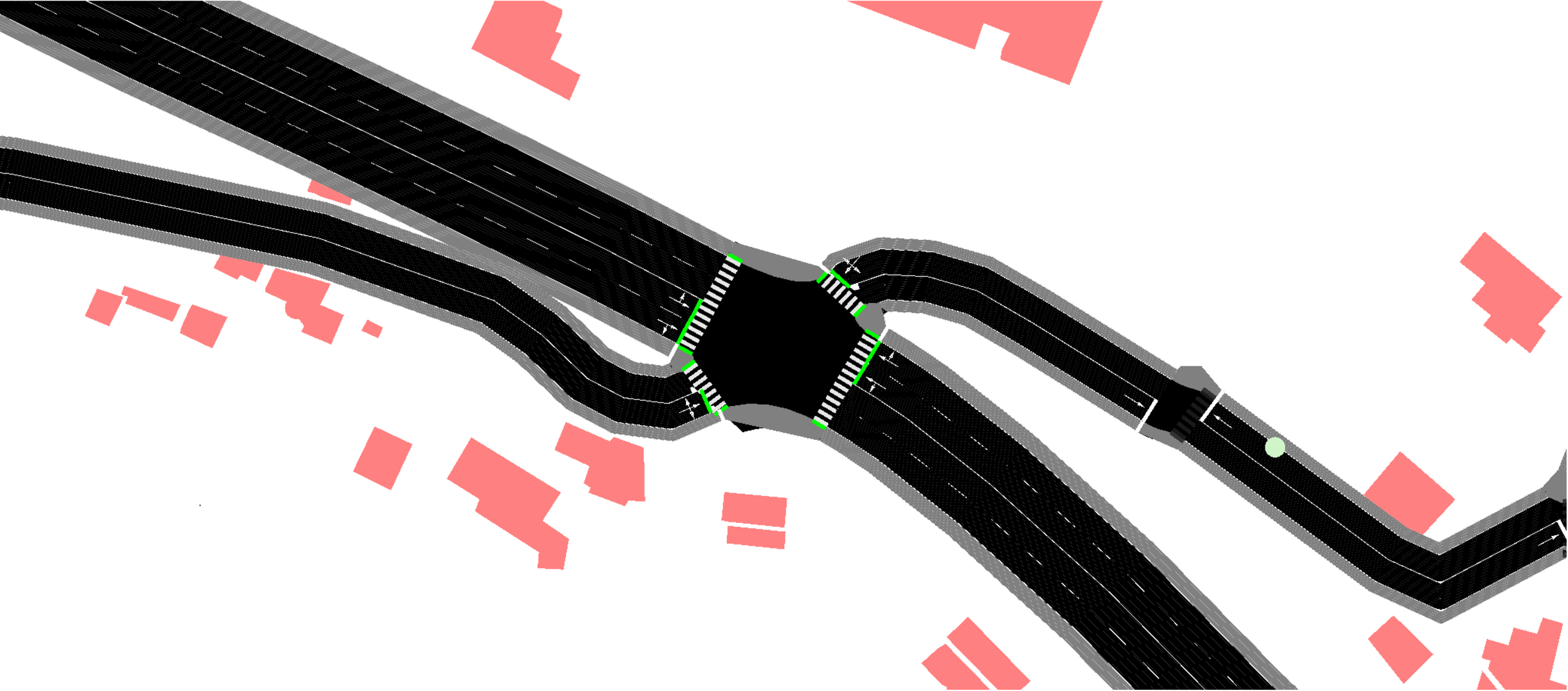}
 \caption{Physical topology of the Luxembourg (left) and Monaco intersections (right). The layout highlights the distinct nature of the presented four-way intersections.}\label{intersections}
\end{figure*}

To evaluate the proposed methodology, we used synthetic mobility traces from pre-built and validated traffic scenarios in the SUMO emulator representing realistic traffic demands and mobility patterns. Specifically, in this work, we extracted the mobility traces from two specific unregulated intersections in Luxembourg \cite{codeca2017luxembourg} and Monaco \cite{MoSTCodeca2017} through the SUMO’s Traffic Control Interface (TraCI) library. The two intersections from different cities enable us to validate the proposed approach under distinct multi-modal traffic demands, vehicle movements, and physical topology. Figure~\ref{intersections} illustrates the layout of the Luxembourg and Monaco intersections.

{\bf Luxembourg Intersection:} It is a four-way intersection with a surface area of 1284\,m$^2$ connecting roads with three lanes. Using TraCI, we collected the vehicle movements at the intersection during a time period including the early morning peak hour, i.e., 6-11\,AM, to create the dataset. The dataset contains 6,132 vehicles, among which 2,035 turned at the intersection while 4,097 were traveling straight. We have also registered 197 vehicle pairs that collided at the intersection.

{\bf Monaco Intersection:} It is a two-lane, four-way intersection with a surface area of 386\,m$^2$ and with a long-stretch highway road on one side connecting the urban driveways at an intersection. Like the Luxembourg intersection, we extracted the vehicle movements during a time period that includes the early morning peak hour, i.e., 7\,AM-12\,PM. The Monaco dataset contains the mobility traces of 5,950 vehicles, with 1,105 turning and 4,845 non-turning at the intersection. In addition, we have recorded 244 vehicle collision pairs at the intersection.   

\begin{table}[b]
\centering
\caption{Collision classifications}
\vspace{-1.5mm}
\label{tab:collisionclassification}
{\begin{tabular}{ccccc}
\hline
								   & Front & Rear & Side  & Corner\\ \hline \noalign{\vskip 0.05in}
{Luxembourg}      & 0 & 0 & 34 & 5\\  \hline
                                   \noalign{\vskip 0.05in}
{Monaco} & 3 & 6 & 34 & 8 \\ \hline
                                   \noalign{\vskip 0.05in}

\end{tabular}}
\end{table}

\textbf{Pre-processing: }For both datasets, the TraCI library facilitates collecting all the features mentioned in Sec.\,\ref{dataset_formulation} at a sampling period of $100$\,ms. Furthermore, to regularize the input features, a feature scaling technique is used. As a result of the scaling technique, all inputs are standardized with zero mean and standard deviation equal to one, allowing the LSTM ED model to converge to a model of high accuracy. The proposed LSTM network takes the last $3$ seconds of data, i.e., $30$ data points, as input to predict the future vehicles’ position for the next $3$ seconds, i.e., $30$ data points. Alternative input and output windows were also tested, with no significant improvement in the obtained results. Additionally, we are using a sliding window technique to share $29$ data points between two consecutive data samples. With this setup, we have $2.4\times 10^6$ and $5.1\times 10^6$ sequences for the Luxembourg and Monaco intersection datasets, respectively. The datasets use $65\%$, $15\%$, and $20\%$ as a split ratio to divide the vehicle trajectories into training, validation, and testing sets, respectively. Among the vehicle collision pairs, the testing set for the Luxembourg and Monaco intersections comprises $39$ and $51$ vehicle collision pairs, respectively. Both the Luxembourg and Monaco testing sets include various collision types. Table\,\ref{tab:collisionclassification} presents a summary of the number of collisions belonging to the rear, front, side, and corner categories. In Table\,\ref{tab:collisionclassification}, we define a ``side collision'' as a scenario where at least one of the two vehicles involved proceeds straight at the intersection, and the other vehicle collides on its side. Conversely, we classify a collision as a ``corner collision'' if both vehicles are turning at the intersection when the crash occurs.

\subsection{Performance Results - Trajectory Prediction}
{\bf Luxembourg Intersection:} 
The proposed LSTM model comprises a single encoder and decoder layer with 300 hidden units each. The decoder layer is followed by a Time Distributed (TD) dense layer with the same hidden units as in the encoder/decoder layers. Finally, the output layers comprise two units representing the predicted (longitudinal and latitudinal) positions. The model is trained with a batch size equal to 48 and a learning rate of 0.0001. Additionally, to avoid overfitting, we have employed the early stopping criteria to end the training when the error is less than 0.001 in the validation set for two epochs. It should be noted that extensive trials with different hidden layers and batch sizes were performed, with the chosen hyperparameters presenting the best performance in terms of prediction errors. Since we are working at an unregulated intersection, the dataset contains a few traffic slowdowns for a short period of time. However, the performance of our trajectory prediction model is not majorly affected by this heterogeneous traffic behavior.
\begin{table}[]
\centering
\caption{Euclidean prediction errors for the Luxembourg dataset}
\vspace{-1.5mm}
\label{tab:SUMO_LIResults}
\begin{tabular}{ccccc}
\hline
								   & Prediction  &  Eucl. & Interaction-Aware & KF Eucl.\\
								   & Time {[}s{]} & Error {[}m{]} & Eucl. Error {[}m{]} & Error {[}m{]}\\ \hline \noalign{\vskip 0.05in}
\multirow{3}{*}{All vehicles}      & $t+1$ & 0.541  & 0.574 & 1.39\\ \cline{2-5} \noalign{\vskip 0.05in}
                                   & $t+2$ & 0.684  & 0.748 & 4.06\\ \cline{2-5} \noalign{\vskip 0.05in}
                                   & $t+3$ & 1.126  & 1.168 & 7.03\\ \hline
                                   \noalign{\vskip 0.05in}
\multirow{3}{*}{\shortstack{Non-turning \\Vehicles}} & $t+1$ & 0.528  & 0.541 & 1.35 \\ \cline{2-5} \noalign{\vskip 0.05in}
                                   & $t+2$ & 0.639   & 0.706 & 3.85 \\ \cline{2-5} \noalign{\vskip 0.05in}
                                   & $t+3$ & 1.032   & 1.112 & 6.56\\ \hline
                                   \noalign{\vskip 0.05in}
\multirow{3}{*}{\shortstack{Turning \\ Vehicles}}   & $t+1$ & 0.563   & 0.631 & 1.48 \\ \cline{2-5} \noalign{\vskip 0.05in}
                                   & $t+2$ & 0.763  & 0.82 & 4.45 \\ \cline{2-5} \noalign{\vskip 0.05in}
                                   & $t+3$ & 1.289  & 1.265 & 7.9 \\ \hline
\end{tabular}
\end{table}
The ED metric is used to evaluate the model’s performance by comparing the predicted positions with the true vehicle positions for the next 3 seconds. As suggested in \cite{deo2018multi}, the performance of Encoder-Decoder LSTM is also compared with a constant speed KF technique, which performs quite well in highways. Furthermore, to evaluate how the proposed technique compares against one leveraging information on multiple surrounding vehicles,  we assess the performance of an LSTM-ED approach that not only considers the preceding vehicle, but also the closest lateral (left and right) and following vehicles.  Hereinafter, we refer to such an approach as Interaction-Aware LSTM-ED.  Table~\ref{tab:SUMO_LIResults} presents the average ED error in predicting the vehicle position  at $t+1$, $t+2$, and $t+3$ seconds. The resulting ED error shows that the proposed LSTM-ED model can predict the future trajectories for the horizon of 3 seconds with minimal deviations from the true trajectories. The proposed approach outperforms also the KF benchmark, showing that, assuming little to no modifications of the vehicles’ behavior at intersections, leads to very large errors both in short and long prediction windows. When considering the best hyperparameter selection for both approaches, the proposed LSTM-ED also performs slightly better than the Interaction-Aware LSTM-ED approach. The obtained results confirm that: (i) given the small error, the performance obtained by the proposed approach is difficult to improve, even including further input data; (ii) in a busy intersection, where a change of lanes is rare, the information on the preceding vehicle is the most relevant one. Specifically, in the datasets at hand, only $8.71\%$ and $2.8\%$ of the vehicles change lane in the surrounding of the intersection in the  Luxembourg and  the Monaco testing set, respectively. For all other vehicles, and for all vehicles changing lanes in all time periods during which they are not changing lanes, the surrounding vehicle information can be considered as noisy input data. Being in most cases unnecessary, surrounding vehicle information can introduce additional complexity to the LSTM model and might hinder the model's ability to generalize well, leading to overfitting on the training data, negatively impacting performance on unseen data.

\begin{figure}[h!]
\begin{center}
\includegraphics[width=1\columnwidth]{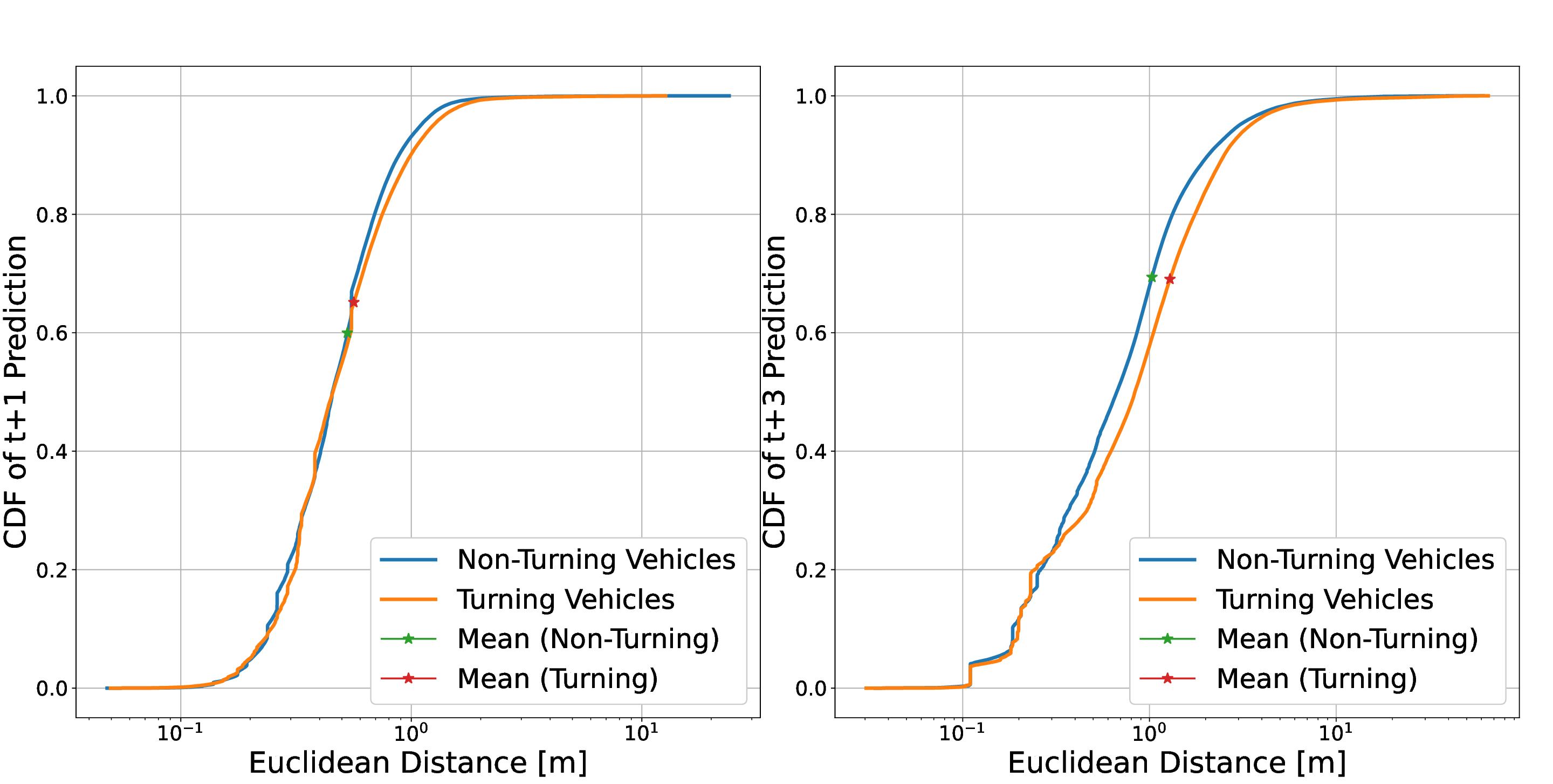}
\caption{Mean error CDF for predicted trajectories at $t+1$ (left) and $t+3$ (right) (Luxembourg dataset).}
\label{CDFSUMO_LI}
\end{center}
\vspace{-5mm}
\end{figure}

Finally, Fig.~\ref{CDFSUMO_LI} presents the CDF of the prediction errors at time $t+1$ and $t+3$ (for the sake of space, the CDF of $t+2$ is not included). The CDF of the prediction errors reveals that not only the average prediction errors but also the distribution of errors are indeed small. Specifically, 91.67\% of the 1-s look ahead predictions exhibit an error of less than 1\,m, while 2-s look ahead presents 96.81\% of errors below 2\,m and 3-s look ahead presents 94.09\% of errors less than 3\,m. 

\begin{figure}[h!]
\begin{center}
\includegraphics[width=1\columnwidth]{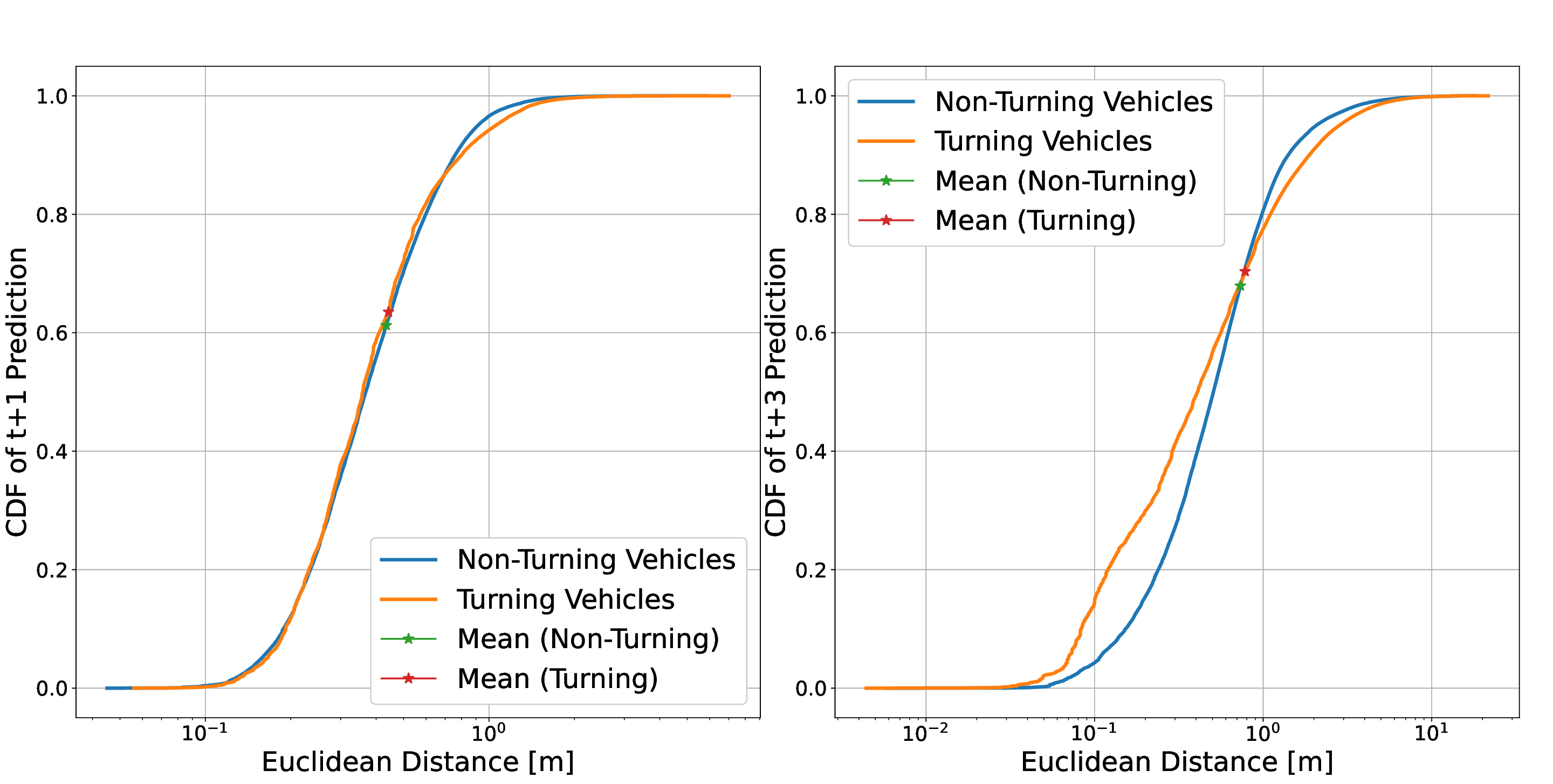}
\caption{Mean error CDF for predicted trajectories at $t+1$ (left) and $t+3$ (right) (Monaco dataset).}
\label{CDFSUMO_MI}
\end{center}
\vspace{-5mm}
\end{figure}

\noindent{\bf Monaco Intersection:}
For the Monaco dataset, the best model utilizes a single-layer encoder and decoder with 300 hidden units each. The TD dense layer contains a hidden layer of size 300 neurons. Like the Luxembourg dataset, we set the output of the final dense layer to 2 units, representing the predicted longitudinal and latitudinal positions. We trained the model with a batch size of 32 and 0.0001 as a learning rate. Again, to avoid overfitting, we employed the early stopping criteria to stop training when the error is less than 0.001 in the validation set for two epochs. As this is also an unregulated intersection, the dataset presents quite a few gridlock situations restricting the vehicles' movements on different occasions in the five hours of collected mobility traces. This traffic scenario poses an additional challenge to the model, as it has to consider congestion while predicting traffic stoppages.
\begin{table}[]
\centering
\caption{Euclidean prediction errors for the Monaco dataset}
\label{tab:SUMO_MIResults}
\begin{tabular}{ccccc}
\hline
								   & Prediction  &  Eucl. & Interaction-Aware & KF Eucl.\\
								   & Time {[}s{]} & Error {[}m{]} & Eucl. Error {[}m{]} & Error {[}m{]}\\ \hline \noalign{\vskip 0.05in}
\multirow{3}{*}{All vehicles}      & $t+1$ & 0.437  & 0.843 & 1.46\\ \cline{2-5} \noalign{\vskip 0.05in}
                                   & $t+2$ & 0.496  & 0.902 & 4.24\\ \cline{2-5} \noalign{\vskip 0.05in}
                                   & $t+3$ & 0.751  & 1.178 & 7.26\\ \hline
                                   \noalign{\vskip 0.05in}
\multirow{3}{*}{\shortstack{Non-turning \\Vehicles}} & $t+1$ & 0.434  & 0.838 & 1.60 \\ \cline{2-5} \noalign{\vskip 0.05in}
                                   & $t+2$ & 0.491   & 0.864 & 4.59 \\ \cline{2-5} \noalign{\vskip 0.05in}
                                   & $t+3$ & 0.733   & 1.099 & 7.74\\ \hline
                                   \noalign{\vskip 0.05in}
\multirow{3}{*}{\shortstack{Turning \\ Vehicles}}   & $t+1$ & 0.442 & 0.85 & 1.17 \\ \cline{2-5} \noalign{\vskip 0.05in}
                                   & $t+2$ & 0.505  & 0.965 & 3.54 \\ \cline{2-5} \noalign{\vskip 0.05in}
                                   & $t+3$ & 0.782  & 1.308 & 6.28 \\ \hline
\end{tabular}
\end{table}
Table~\ref{tab:SUMO_MIResults} presents the average ED error between actual and predicted trajectories for a 3-second horizon of the proposed LSTM-ED approach, of a constant speed KF \cite{deo2018multi}, and of the Interaction-Aware LSTM-ED benchmark introduced above. Even with the additional complexity of increased traffic congestion, the results show that the model is able to achieve sub-meter prediction errors in all prediction windows and outperform both benchmark approaches. Figure~\ref{CDFSUMO_MI} presents the CDF of the prediction errors for $t+1$ and $t+3$ seconds look ahead. In all cases, the CDF shows that the distribution of errors is small and similar to the Luxembourg dataset. Specifically, 95.41\% of 1-s look ahead predictions exhibit less than 1\,m error, 98.25\% of 2-s look ahead (omitted for the sake of space) have prediction error less than 2\,m, and 96.49\% of 3-s look ahead predictions maintain less than 3\,m error. 

The results from both datasets indicate that the proposed approach can excel in diverse scenarios and mobility conditions. Several other comparisons (also including real-world datasets) have been reported in our previous work \cite{Dinesh21}. Specifically, the proposed approach was found to perform better than (i) a Vanilla LSTM model, 
(ii) a Stacked LSTM-ED model, and (iii) a Bidirectional LSTM model. These results have been omitted here 
due to space limitations.

\subsection{Performance Results - Trajectory Uncertainty Estimation}
A prediction interval estimates the uncertainty of a point prediction with probabilistic upper and lower bounds on the output variable. For this task, as discussed earlier in Sec. \ref{sec:trajectory_estimation}, an LSTM-ED model trained to minimize the asymmetrically weighted sum of absolute errors is used. With the objective of characterizing the tails of the trajectory prediction's distribution, for this work, we have utilized the $0.9$-quantile as the upper bound $u$ and the $0.1$-quantile as the lower bound $l$. Hence, the geographical regions individuated by the proposed trajectory uncertainty estimation are expected to contain in total $80\%$ of the vehicle’s future positions. Unlike the point prediction model, the deep quantile regression approach uses two models to separately predict the vehicle’s longitudinal and latitudinal positions. This solution is motivated by the obtained performance, which approximates closely the expected number of the vehicles' future locations included in the prediction intervals. Further, as in point predictions, the models utilize early stopping criteria controlling the training epochs.

For both the Luxembourg and Monaco datasets, the best LSTM-ED model comprises a single encoder/decoder layer with $320$ LSTM units. The decoder output is linked to a TD dense layer using $128$ neurons, while the final output dense layer presents two values, one for the lower and one for the upper quantile estimation. The model uses $48$ as a batch size with $0.0001$ as a learning rate to minimize the quantile loss function for a specific number of epochs.
\begin{table*}[]
\centering
\caption{Trajectory uncertainty estimation: The percentage of true vehicle trajectories captured by the $0.9$ and $0.1$ quantiles and the percentage of true vehicles in-between the quantiles.}
\vspace{-1.5mm}
\label{tab:QuantileResults}
\begin{tabular}{cccccccc}
\hline
\multirow{4}{*}{Dataset} & \multirow{4}{*}{Prediction Time {[}s{]}} & \multicolumn{6}{c}{Quantile Estimation {[}\%{]}} \\ \cline{3-8} \noalign{\vskip 0.05in}
                                   &     & \multicolumn{2}{c}{$0.9$} & \multicolumn{2}{c}{$0.1$} & \multicolumn{2}{c}{In-between}  \\ \cline{3-8}
                                   \noalign{\vskip 0.05in}
                                  & & $\phi$ & $\lambda$ & $\phi$ & $\lambda$ & $\phi$ & $\lambda$  \\ \cline{3-8} \hline 
                                  \noalign{\vskip 0.05in}
\multirow{3}{*}{Luxembourg}      & $t+1$ & 92.47 & 95.29 & 8.58 & 8.38 & 83.89 & 86.91 \\ \cline{3-8}                                        \noalign{\vskip 0.05in}
                                   & $t+2$ & 90.58 & 93.38 & 5.14 & 7.09 & 85.44 & 86.29 \\ \cline{3-8} 
                                   \noalign{\vskip 0.05in}
                                   & $t+3$ & 89.94 & 92.56 & 5.23 & 7.52 & 84.71 & 85.04  \\ \hline
                                   \noalign{\vskip 0.05in}
\multirow{3}{*}{Monaco}          & $t+1$ & 99.60 & 99.04 & 1.33 & 7.37 & 98.27 & 91.67 \\ \cline{3-8} 
                                    \noalign{\vskip 0.05in}
                                   & $t+2$ & 98.00 & 98.70 & 2.16 & 7.42 & 95.84 & 91.28 \\ \cline{3-8} 
                                   \noalign{\vskip 0.05in}
                                   & $t+3$ & 96.00 & 97.67 & 4.42 & 7.69 & 91.57 & 89.98 \\ \hline
\end{tabular}
\end{table*}

Table~\ref{tab:QuantileResults} presents the deep quantile regression results for the Luxembourg and Monaco datasets regarding the percentage of the vehicle's true locations captured by the computed quantiles for $t+1$, $t+2$, and $t+3$ prediction horizons on the latitudinal ($\lambda$) and longitudinal ($\phi$) directions. The $0.9$ and $0.1$ quantiles, i.e., upper and lower bounds, present the ratio of true vehicle positions below the predicted output, while the ``in-between'' column presents the percentage of true vehicle locations in-between the two quantiles. From the results of both datasets, we can determine that the $0.9$ and $0.1$ quantiles included, respectively, slightly more and slightly less samples than expected. Notably, even though the tested samples are unseen in the testing dataset, the number of true vehicle locations included between the obtained quantiles approximate closely its expected value. Indeed, $89.24\%$ of vehicle trajectories are between the two quantiles’ output on average, as compared to the expected $80\%$.

\subsection{Performance Results - Collision Detection}
To identify any collision at the intersection, our detection algorithm hosted at the IM utilizes an RFC exploiting as input, at any time $t$, both the vehicles' predicted trajectories and the vehicles' location uncertainty estimations for the following 3 seconds. Hence, the objective of our approach is to detect a collision 3 seconds before it actually happens, i.e., as soon as the predicted data includes also the moment of collision. Even though, due to the output provided by SUMO, all pairs of colliding vehicles can be easily identified, it is possible that a collision is not easily identifiable when looking only at the true locations of the vehicles. Indeed, given the fact that the location of the vehicles corresponds to the vehicles' front bumper, it is possible that two vehicles will collide even though the distance between their true locations is always greater than zero. This may happen, e.g., if the front bumper of a vehicle collides with the rear bumper of another vehicle. In order to make all collisions uniform from the RFC's perspective, during training an input sequence concerning a pair of colliding vehicles $i$ and $k$ is manually assigned to the ``collision'' class if one the following three events occurs:
\begin{itemize}
    \item within three seconds, vehicles $i$ and $k$ collide;
    \item $d_{ik,t+j}$ is less than $d_c$, with $d_c$ being the $0.9$-quantile of the minimum distance between the true locations of training colliding vehicles' pairs;
    \item $E[d^2_{ik,t+j}]$ is less than $d^2_c$, with $d^2_c$ being the $0.9$-quantile of the minimum squared distance between the true locations of training colliding vehicles' pairs.
\end{itemize}

For the Luxembourg and Monaco datasets, $d_c$ is set to 4.87\,m and 3.51\,m, while $d^2_c$ is specified as 23.73\,$m^2$ and 12.34\,$m^2$, respectively. As detailed later, this choice is driven by the fact that the ED and the expected squared ED between the estimated vehicles' locations play a crucial role in collision detection. Hence, a diverse characterization of such metrics during training would produce unbalanced results, depending on the type of collision at hand. Furthermore, during inference, a DENM is sent to the vehicles expected to collide if a collision is predicted for three consecutive time-steps, i.e., for a time interval of 300\,ms. This choice for the sporadicity filter represents the numerical optimal solution to restrict the number of False Positives (FPs), while detecting all pairs of vehicles on a collision course. The performance of the obtained solution is compared to two benchmark solutions:
\begin{itemize}
    \item an approach (named ``Relative Distance'') based only on the point estimates obtained in Sec. \ref{sec:trajectory_estimation}, that transmits an alarm to the corresponding pair of vehicles, as soon as the ED between the predicted locations of vehicles is below $d_c$;
    \item a state-of-the-art solution (named ``Cooperative Intersection - Collision Warning System (CI-CWS)'') \cite{marco_CA} that uses the vehicles' velocity, acceleration, and position to predict their collision time by solving a fourth-degree polynomial approximation and that consequently transmits an alarm to the vehicles in danger.
\end{itemize}

\begin{figure}[h!]
\begin{center}
\includegraphics[scale=1]{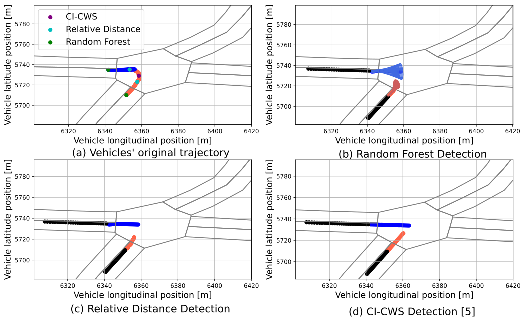}
\caption{Collision detection analysis for the Luxembourg intersection.}
\label{fig:CD_Analysis_Li}
\end{center}
\vspace{-5mm}
\end{figure}

\begin{figure}[h!]
\begin{center}
\includegraphics[scale=1]{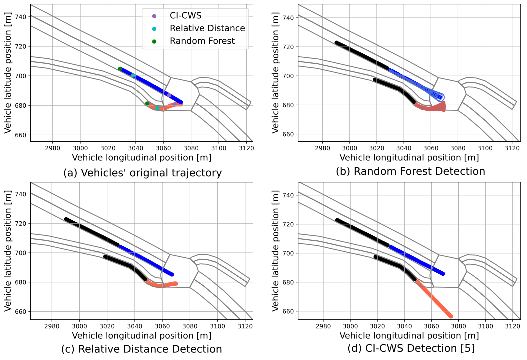}
\caption{Collision detection analysis for the Monaco intersection.}
\label{fig:CD_Analysis_Mo}
\end{center}
\vspace{-5mm}
\end{figure}

\subsubsection{Collision Detection Examples}
Before analyzing the performance of the proposed approach in all the collisions present in the testing set, we first choose two collisions that exemplify the role of the uncertainty estimation in the proposed framework. Figures \ref{fig:CD_Analysis_Li} and \ref{fig:CD_Analysis_Mo} represent the two selected collisions for the Luxembourg and the Monaco intersection, respectively. Both figures refer to the moment at which our proposed methodology, based on the RFC, identifies a possible dangerous situation and sends a DENM message to the two vehicles involved. The top-left image of each figure illustrates the trajectory traveled in the dataset by the two colliding vehicles, starting from the moment where the RFC triggers the alarm (detection and collision are interleaved by 8.8 seconds in the Luxembourg intersection, and by 3.9 seconds in the Monaco intersection). Herein, green, blue, and purple, mark the location of the vehicles when the RFC, the detection performed exploiting relative distance only, and the approach in \cite{marco_CA}, respectively, detect the imminent danger. The remaining three images, instead, depict the 3-second trajectory predictions used by our approach, by the detection performed exploiting relative distance only, and by the approach in \cite{marco_CA}, to predict a possible collision. Furthermore, the image referring to our approach, also reports the uncertainty estimation associated with each of the future locations predicted for the vehicles involved. Such uncertainty is represented graphically as an ellipse, with major and minor axes equal to the prediction intervals obtained with the LSTM model using a quantile loss function. In both Fig.\,\ref{fig:CD_Analysis_Li}(a) and Fig.\,\ref{fig:CD_Analysis_Mo}(a), our proposed collision detection technique reacts earlier than any other approach to the imminent danger. Even though the trajectory predictions obtained by the LSTM are not showcasing a possible collision (the front bumpers are still at least $8$\,m away in both intersections), the associated prediction intervals are large, indicating that multiple possible driver maneuvers are available in both cases (as depicted in Fig.\,\ref{fig:CD_Analysis_Li}(b) and Fig.\,\ref{fig:CD_Analysis_Mo}(b)). Based on the knowledge of dangerous situations present in the training set of the Monaco and Luxembourg intersection, the RFC  used in our approach identifies as dangerous the inputs given, i.e., based on the speed of the vehicles, the predicted trajectories, and the associated uncertainty, the RFC correctly detects that the two vehicles may collide. Hence, our proposed approach can react to imminent dangers even if the vehicles trajectories appear to be safe, allowing for a smooth collision avoidance, even in the presence of human drivers. Simple trajectory predictions, as the ones used in \cite{marco_CA} and showcased in Fig.\,\ref{fig:CD_Analysis_Li}(d) and Fig.\,\ref{fig:CD_Analysis_Mo}(d), fail to provide a reliable predicted location for vehicles at intersections, and, as a result, detect  possible collisions very late.

\subsubsection{Statistical Comparative Analysis}

\begin{table}[]
\centering
\caption{Comparison of collision detection methods (Luxembourg dataset).}
\vspace{-1.5mm}
\label{tab:LI_CD}
\begin{tabular}{cccc}
\hline
\multirow{2}{*}{}  & \multicolumn{3}{c}{Collision Detection Method} \\ \cline{2-4} 
                            \noalign{\vskip 0.05in}
                          & Random Forest & Relative Distance & CI-CWS \\ \hline
                          \noalign{\vskip 0.05in}
                    False Positive   & 30 & 18 & 186    \\ \cline{1-4}  \noalign{\vskip 0.05in}
                    False Negative  & 0  & 0 & 1    \\ \cline{1-4}  \noalign{\vskip 0.05in}
                    True Positive   & 39 & 39 & 38    \\ \hline
\end{tabular}
\end{table}

\begin{table}[]
\centering
\caption{Comparison of collision detection methods (Monaco dataset).}
\vspace{-1.5mm}
\label{tab:MI_CD}
\begin{tabular}{cccc}
\hline
\multirow{2}{*}{}  & \multicolumn{3}{c}{Collision Detection Method} \\ \cline{2-4} 
                            \noalign{\vskip 0.05in}
                           & Random Forest & Relative Distance & CI-CWS \\ \hline 
                          \noalign{\vskip 0.05in}
                    False Positive   & 34 & 24 & 419    \\ \cline{1-4} \noalign{\vskip 0.05in}
                    False Negative   & 0  & 2 & 2    \\ \cline{1-4} \noalign{\vskip 0.05in}
                    True Positive   & 51 & 49 & 49    \\ \hline
\end{tabular}
\end{table}

Tables~\ref{tab:LI_CD} and \ref{tab:MI_CD} summarize the predicted collision detection results for the Luxembourg and Monaco intersections, respectively. We have performed a collision check for each pair of vehicles that came as close as 50 meters, i.e., 45,674 (Luxembourg) and 69,712 (Monaco) collision checks for the two intersections analyzed. For the Luxembourg intersection, our approach, named herein ``Random Forest'', and the ``Relative Distance'' collision detection method were able to identify all the collisions that occurred in the testing set, i.e., zero False Negatives were obtained, irrespective of the type of collision.  Notwithstanding, for the Monaco intersection, the ``Relative Distance'' approach was not able to detect all collisions with the $0.9$-quantile as a distance threshold while ``Random Forest'' was able to identify all of them.
Contrary to the introduced methods, the CI-CWS algorithm \cite{marco_CA} is not able to detect all collisions at either intersection. This is due to the less accurate trajectory prediction used, which is also reflected by a much higher number of FPs in the simulated time frames.
\begin{table}[]
\centering
\caption{Distribution of distance range for FP detections (Luxembourg dataset).}
\vspace{-1.5mm}
\label{tab:LI_FPDistro}
\begin{tabular}{cccc}
\hline
\multirow{2}{*}{\shortstack[c]{Vehicle body \\ distance range {[}m{]}}}  & \multicolumn{3}{c}{Number of occurrences} \\ \cline{2-4} \noalign{\vskip 0.05in}
                         & Random Forest & Relative Distance & CI-CWS \\ \hline \noalign{\vskip 0.05in}
                   $0-5$  & 12 & 16 & 103    \\ \cline{1-4}  \noalign{\vskip 0.05in}
                   $5-10$  & 10 & 1 & 53    \\ \cline{1-4} \noalign{\vskip 0.05in}
                   $10-15$ & 4  & 1  & 17    \\ \cline{1-4}  \noalign{\vskip 0.05in}
                   $\geq 15$  & 4 & 0 & 13    \\ \hline
\end{tabular}
\end{table}

\begin{table}[]
\centering
\caption{Distribution of distance range for FP detections (Monaco dataset).}
\vspace{-1.5mm}
\label{tab:MI_FPDistro}
\begin{tabular}{cccc}
\hline
\multirow{2}{*}{\shortstack[c]{Vehicle body \\ distance range {[}m{]}}}  & \multicolumn{3}{c}{Number of occurrences} \\ \cline{2-4} \noalign{\vskip 0.05in}
                         & Random Forest & Relative Distance & CI-CWS \\ \hline \noalign{\vskip 0.05in}
                   $0-5$  & 31 & 23 & 319    \\ \cline{1-4} \noalign{\vskip 0.05in}
                   $5-10$   & 2 & 0 & 87    \\ \cline{1-4} \noalign{\vskip 0.05in}
                   $10-15$  & 1 & 1 & 12    \\ \cline{1-4} \noalign{\vskip 0.05in}
                   $\geq 15$  & 0 & 0 & 1    \\ \hline
\end{tabular}
\end{table}

Furthermore, Tables~\ref{tab:LI_FPDistro} and \ref{tab:MI_FPDistro} show the FP count distribution with respect to the minimum distance between vehicles. For this set of results, the actual minimum distance between the vehicles' bodies is reported, rather than the distance between the true location of the vehicles. Almost all FPs are related to vehicles that, in the simulation, are actually getting very close to each other. Among the Random Forest model's FPs, at least 66\% of the vehicles are in a critical region with less than a $5$-meter distance between them. In some cases, especially in the Luxembourg dataset, our proposed Random Forest approach detects possible collisions, even though the true minimum distance between them never goes below $15$ m. In three of the four such predicted collisions, one of the two drivers in the simulation applied the brakes to evade the collision, leading to a longer distance for the FP. As a confirmation that the trajectories used by the CI-CWS algorithm are less accurate than the ones used by our approach, Tables~\ref{tab:LI_FPDistro} and \ref{tab:MI_FPDistro} also show that, in many cases, the alarms raised with the considered benchmark approach concerns vehicles whose distance never drops below 15\,m.

\begin{figure}[h!]
\begin{center}
\includegraphics[width=0.85\columnwidth]{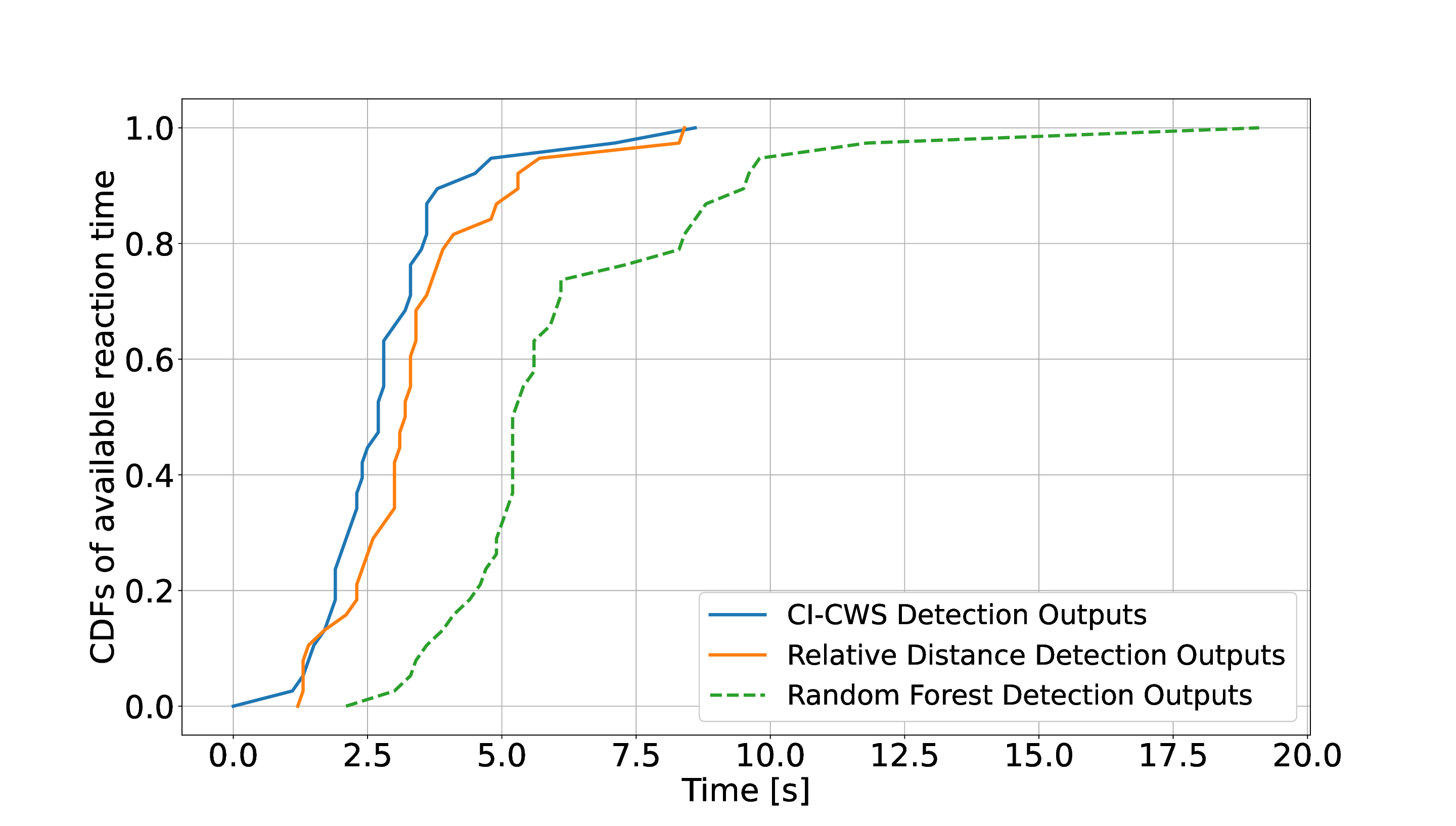}
\caption{CDF of the available reaction time (Luxembourg dataset).}
\label{CDF_LI_RT}
\end{center}
\vspace{-5mm}
\end{figure}

\begin{figure}[h!]
\begin{center}
\includegraphics[width=0.85\columnwidth]{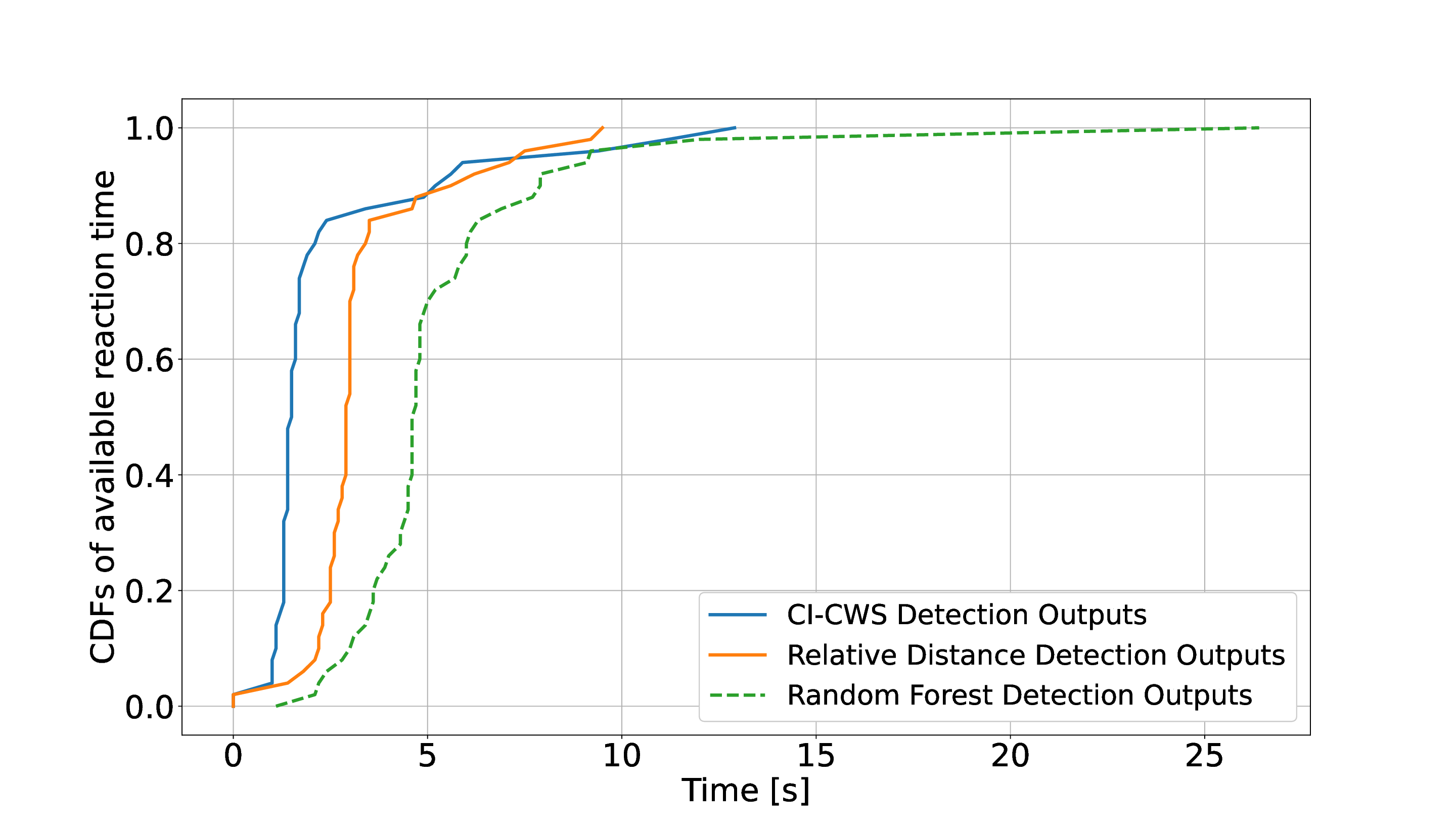}
\caption{CDF of the available reaction time (Monaco dataset).}
\label{CDF_MI_RT}
\end{center}
\vspace{-5mm}
\end{figure}

The effect of accounting for uncertainty in the detection mechanism is clear when a critical metric for avoiding a collision is considered, i.e., the time between the detection of a collision and the time of the collision, which we refer to as ``available reaction time'' and denote with $T_c$ in Sec.\,\ref{sec:collision_avoidance}. Figures~\ref{CDF_LI_RT} and~\ref{CDF_MI_RT} showcase the performance of the Random Forest and Relative Distance approaches through the CDF of the available reaction time. For the Luxembourg intersection, Fig.~\ref{CDF_LI_RT} shows that the Relative Distance approach allows as low as $1.2$\,s to react to a collision, which is insufficient for any driver to perform a corrective action. However, with the Random Forest model, we can detect all the collisions $2$\,s before they occur, and the distribution of the available reaction time is better compared to the Relative Distance method for any time instance. Similarly, for the Monaco intersection, Fig.~\ref{CDF_MI_RT} indicates that the Random Forest model detected all but one collisions before the $2$-second mark, outperforming the Relative Distance method, which could not identify all the collisions.

It is worth noting that, in this dataset, our Random Forest model detects one  collision (out of $69,712$ cases) $20$\,s or more earlier than the actual moment of collision. This happens at the time instant when two vehicles are queued at the intersection, and get on a collision course as they accelerate to cross the intersection. In the simulation, thanks to a corrective action, i.e., braking, the vehicles are initially able to avoid the collision. Nevertheless, when they actually restart a few seconds later, they in fact collide. The Random Forest approach can detect both the dangerous situations described above. Overall, it can be concluded that the Random Forest detection method is efficient and outperforms any of the presented baseline techniques. In addition to detecting all collisions, the proposed Random Forest approach allows for a larger available reaction time for the drivers to take corrective action. This holds true for the CI-CWS algorithm as well, which also yields a much higher rate of FPs.

\begin{figure}[h!]
\begin{center}
\includegraphics[width=1\columnwidth]{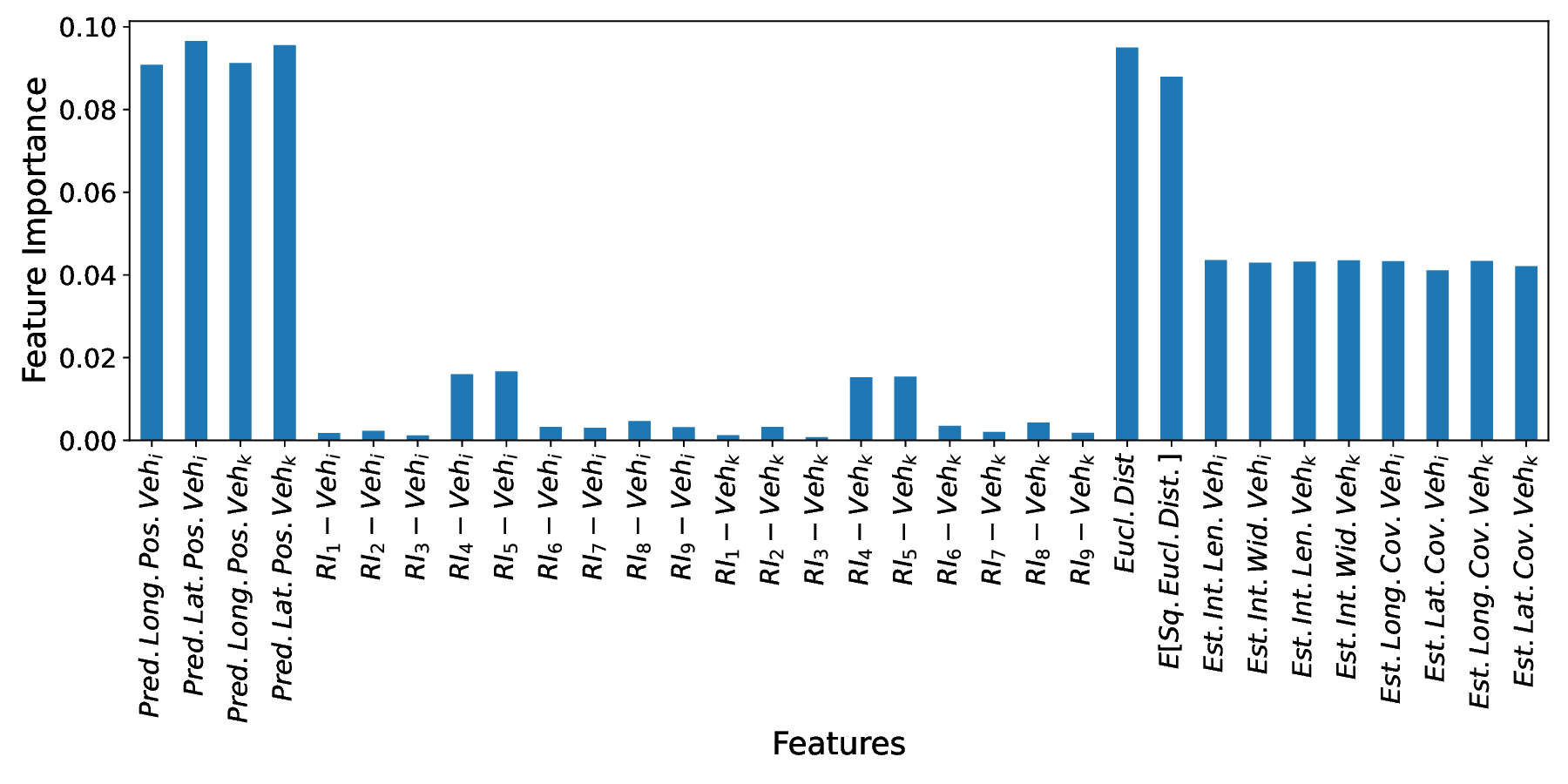}
\caption{Input feature importance for the RFC (Luxembourg dataset).}
\label{LI_FI}
\end{center}
\vspace{-5mm}
\end{figure}

\begin{figure}[h!]
\begin{center}
\includegraphics[width=1\columnwidth]{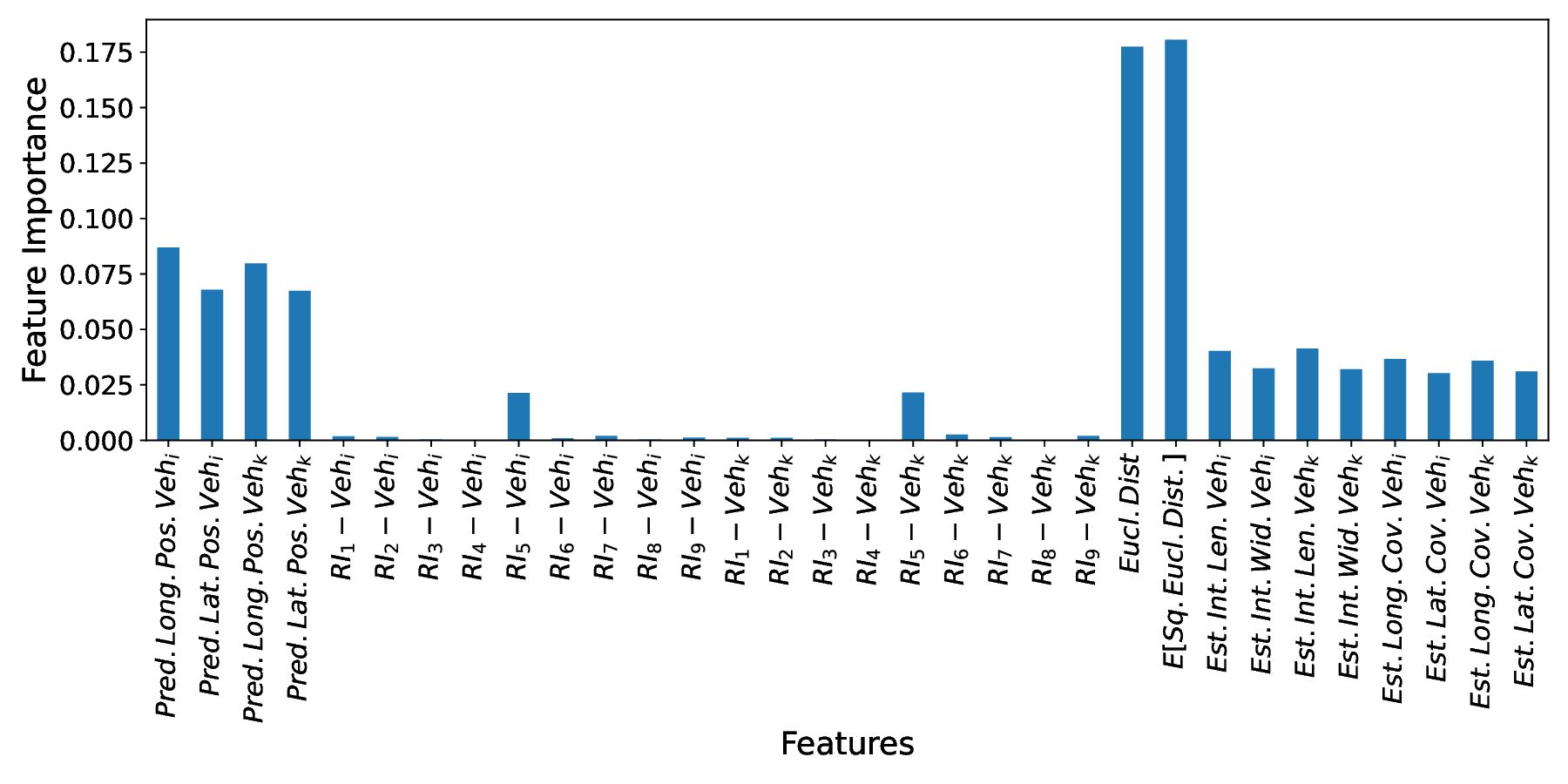}
\caption{Input feature importance for the RFC (Monaco dataset).}
\label{MI_FI}
\end{center}
\vspace{-5mm}
\end{figure}

Finally, Figs.~\ref{LI_FI} and~\ref{MI_FI} demonstrate the importance of the input features in detecting collisions. As discussed in Sec.\,\ref{sec:decision_tree}, the Random Forest model comprises a set of decision trees in which each decision tree is built by splitting data into subsets based on the GI. The feature importance specifies the average reduction of the GI by each feature across all trees in the Random Forest model. Amongst all features of the Random Forest model, the predicted distance and the average square distance metrics play a crucial role in detecting collisions for both datasets. This result (i) justifies the introduced manual labeling mentioned earlier; and (ii) indicates that the Random Forest model exploits both the trajectory prediction and uncertainty estimation in order to detect collisions.

\subsection{Performance Results - Collision Avoidance}
\label{subsec:collision_avoidance}
As mentioned earlier, the performance of the proposed approaches in what regards collision avoidance mainly depends on the available reaction time. With the proposed approach, we can identify the collision timely compared to other approaches. Nevertheless, to assess if such detection is enough to avoid a collision, we must consider the latency pertaining to different components, such as communication, in-vehicle data processing, and driver reaction time (Sec.\,\ref{sec:collision_avoidance}). Through post-processing, we tested the collision avoidance application by enforcing a continuous braking action for both vehicles till they came to a halt. For the braking action, we have used normal (4.5\,m/s$^2$) and critical (9\,m/s$^2$) deceleration rates to stop the vehicles. In addition, we have repeated the process for 20 simulation trials with randomly distributed latency values for each component. Specifically, (i) for the communication end-to-end latency component, $T_x$, we used a beta distribution, $B(\alpha=2.0, \beta=5.0)$, scaled between $2.4$ (minimum) and $18$ (maximum) ms \cite{Ma_5GV2X}; (ii) for the in-vehicle processing data upon receiving the DENM, $T_p$, a constant latency of $400$\,ms is considered \cite{marco_CA}; while (iii) for the driver reaction time,  $T_r$, we have used a truncated normal distribution, $\mathcal{N}(\mu=680, \sigma=145, a=-1.24, b=1.52)$, for human-driven vehicles and $T_r=0$ for automated vehicles \cite{pawel_DRT}. The processing time, $T_d$, was determined through extended simulations. The LSTM models' computation time, for both vehicle trajectory prediction and uncertainty estimation, was measured to be approximately $11$ ms. Likewise, the RFC model, which checks whether a pair of vehicles is on a collision course, was evaluated to take around $12$ ms. In both cases, computation time is almost deterministic, with no deviation from the average. Hence, further assuming that collision tests are performed in parallel, $T_d$ is set to $23$ ms.

\begin{table}[]
\centering
\caption{Vehicle's collision speed with and without corrective action (Luxembourg dataset).}
\label{tab:reduced_speed_LI}
\begin{tabular}{cccc}
\hline
{\shortstack[c]{Deceleration \hspace{-4mm}\\ Rate}}  & {\shortstack[c]{Vehicle \hspace{-4mm}\\ Pair}}  & {\shortstack[c]{Vehicle Speed \\ (w/o corrective action) \hspace{-4mm} \\ {[}m/s{]}}} & {\shortstack[c]{Vehicle Speed \\ (with corrective action) \hspace{-4mm} \\ {[}m/s{]}}}\\ \hline \noalign{\vskip 0.05in}
Normal
                  & $1$ & {\shortstack[c]{$Veh_1 = 12.09$ \\ $Veh_2 = 6.26$}} & {\shortstack[c]{$Veh_1 = 5.56$ \\ $Veh_2 = 4.53$}}   \\ 
\hline
\end{tabular}
\end{table}

\begin{table}[]
\centering
\caption{Vehicle's collision speed with and without corrective action (Monaco dataset).}
\label{tab:reduced_speed_MI}
\begin{tabular}{cccc}
\hline
{\shortstack[c]{Deceleration \hspace{-4mm}\\ Rate}}  & {\shortstack[c]{Vehicle \hspace{-4mm}\\ Pair}}  & {\shortstack[c]{Vehicle Speed \\ (w/o corrective action) \hspace{-4mm} \\ {[}m/s{]}}} & {\shortstack[c]{Vehicle Speed \\ (with corrective action) \hspace{-4mm} \\ {[}m/s{]}}}\\ \hline \noalign{\vskip 0.05in}
Critical
                  & $1$ & {\shortstack[c]{$Veh_1 = 12.16$ \\ $Veh_2 = 8.20$}} & {\shortstack[c]{$Veh_1 = 11.48$ \\ $Veh_2 = 7.86$}}   \\        \cline{1-4} \noalign{\vskip 0.05in} \\
Normal
                  & $1$ & {\shortstack[c]{$Veh_1 = 12.16$ \\ $Veh_2 = 8.20$}} & {\shortstack[c]{$Veh_1 = 11.82$ \\ $Veh_2 = 8.20$}}   \\        
                  \hline
\end{tabular}
\end{table}

At both intersections, all automated vehicles involved in a collision were able to safely reach a halt position with critical deceleration. In the case of human-driven vehicles, as shown in Table\,\ref{tab:reduced_speed_LI} for Luxembourg (with normal deceleration rate) and Table\,\ref{tab:reduced_speed_MI} for Monaco (considering both critical and normal deceleration rates), we successfully avoided all collisions but one (a side collision at each intersection). Notably, the immediate danger in both cases stemmed from an abrupt change in the driver's behavior, specifically hard braking, rather than from the type of collision itself. Table\,\ref{tab:reduced_speed_LI} shows that in all cases our detection algorithm helps to reduce considerably the collision intensity. For example, the vehicle speeds are sensibly reduced when collisions happen by applying a normal deceleration rate, i.e., on average the speed was reduced by $23.1\%$ with respect to the case when vehicles collide with no corrective action taken (i.e., without the assistance of our collision detection algorithm).

\section{Conclusions and Future Work\label{sec:concl}}
In this work, we addressed the critical problem of collision detection in complex road segments, e.g., at intersections, and presented a novel framework that effectively leverages an Intersection Manager hosted at the MEC of a 5G network, as well as V2I and I2I communications, to collect and process the relevant data. Due to its advantageous location, the Intersection Manager can obtain a holistic view of the system and, exploiting LSTM RNNs, it can provide accurate trajectory predictions and an estimation of the prediction uncertainty. Exploiting for the first time a ML-aided network-assisted collision detection methodology yielded high-quality performance, i.e., a very-low number of unnecessary FPs and an ability to detect dangerous situations well in advance, even under real-world conditions. Performance comparisons against state-of-the-art solutions and  point-estimates show that the exploitation of a holistic view of the system and the inclusion of prediction intervals in the collision detection application are the critical factors for achieving the showcased results.

Future research will focus on determining similarities among different use case scenarios that will lead to retaining the main information of the obtained trajectory prediction and uncertainty trajectory estimation models, so as to reduce training time. Additionally, to reduce the complexity of multimodal frameworks and to further improve 
collision avoidance in presence of human-driven vehicles, our trajectory prediction and uncertainty 
estimation approach will also be applied to a selected number of possible alternative driver maneuvers 
(e.g., turning right or left, proceeding straight). Finally, the use of federated learning to preserve 
the vehicles' private information will be explored.

\bibliographystyle{IEEEtran}
% \bibliography{IEEEabrv,references}

\vspace{-0.5in}

\begin{IEEEbiography}[{\includegraphics[width=1in,height=1.25in,clip,keepaspectratio]{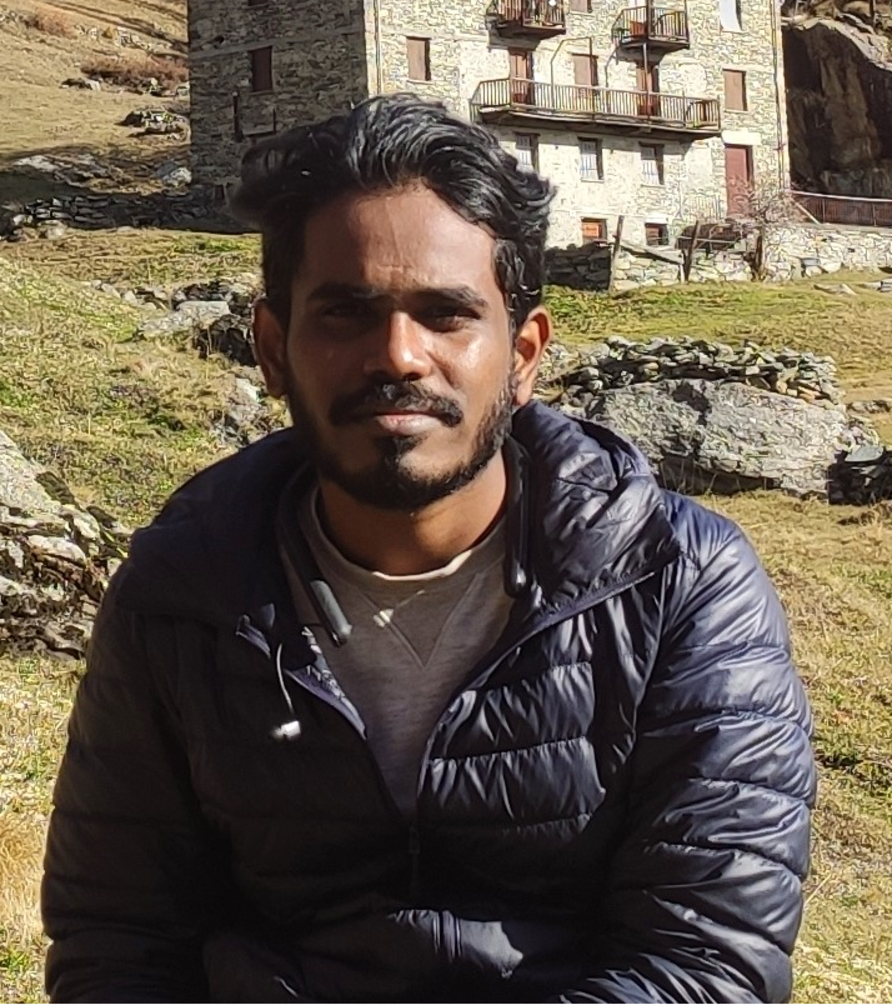}}]{Dinesh Cyril Selvaraj} holds an M.Sc. in ICT for Smart Societies from Politecnico di Torino. He is currently pursuing a Ph.D. degree in the Department of Electronics and Telecommunications, Politecnico di Torino, Italy. His research interests include intelligent transportation systems, V2X technologies, and machine learning-related applications to envision situation-aware vehicles.
\end{IEEEbiography}

\begin{IEEEbiography}[{\includegraphics[width=1in,height=1.25in,clip,keepaspectratio]{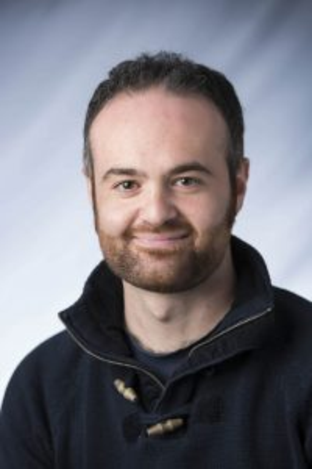}}]{Christian Vitale} holds a B.S. and a M.Sc., from Universit\`{a} di Pisa and a Ph.D. in Telecommunication Engineering from Universidad Carlos III de Madrid. He is a Research Associate at the KIOS Research and Innovation Center of Excellence at the University of Cyprus, where he received an Individual Widening Marie-Curie Fellowship. His interests mainly focus on analytical modeling of complex systems, e.g., wireless networks and intelligent transportation systems, design of mechanisms improving network efficiency, and algorithms for guaranteeing QoS to vertical applications in cellular environments.
\end{IEEEbiography}

\begin{IEEEbiography}[{\includegraphics[width=1in,height=1.25in,clip,keepaspectratio]{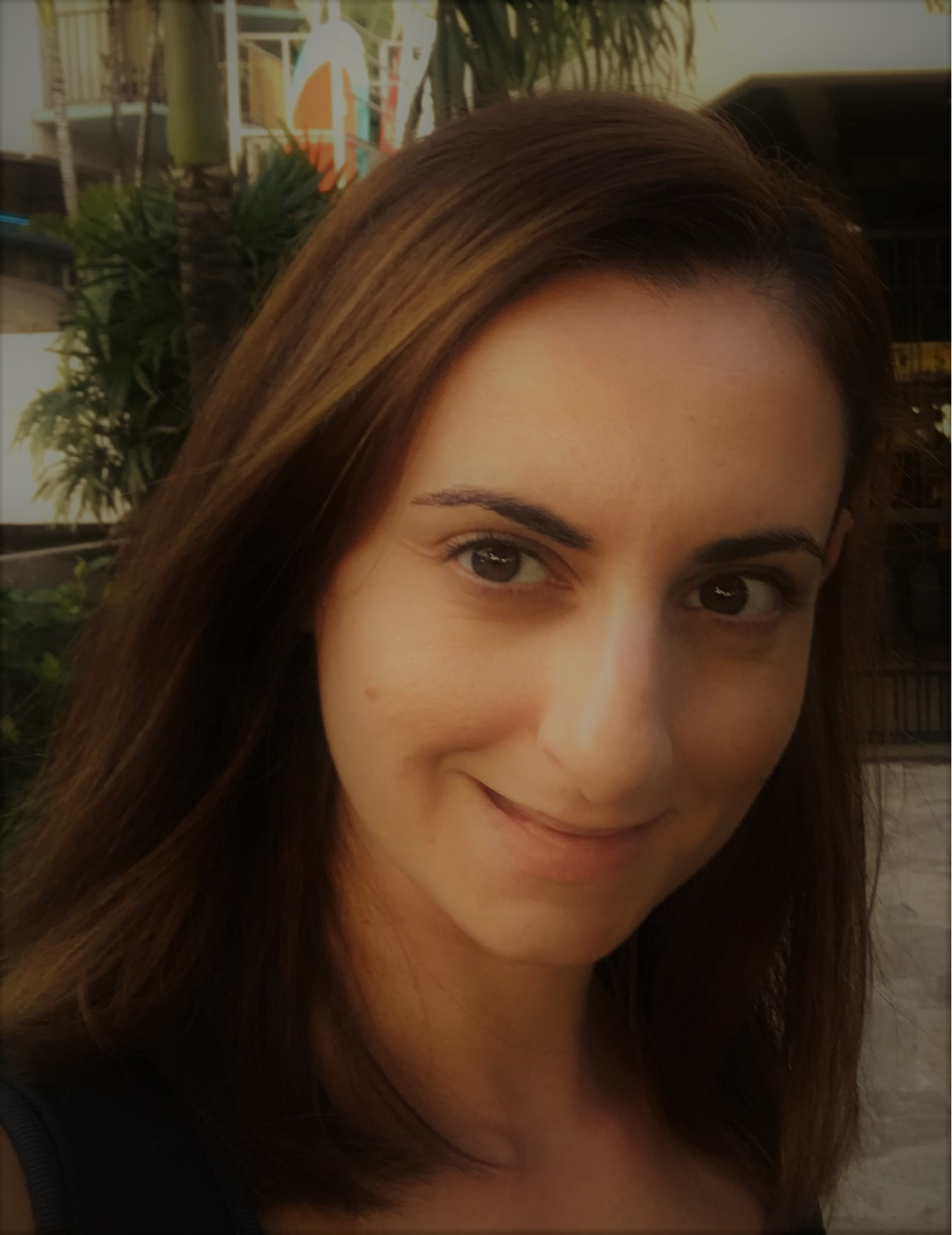}}]{Tania Panayiotou} received the Diploma degree in Computer Engineering and Informatics from the University of Patras, Patras, Greece, in 2005 and the Ph.D. degree in Computer Engineering from the University of Cyprus (UCY) in 2013. She is currently a Research Associate with the KIOS Research and Innovation Center of Excellence, UCY. She has authored more than 40 articles, conference papers, and book chapters. Her research interests include optical networks and transportation networks.
\end{IEEEbiography}

\begin{IEEEbiography}[{\includegraphics[width=1in,height=1.25in,clip,keepaspectratio]{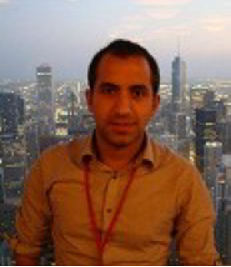}}]{Panayiotis Kolios} received the B.Eng. (2008) and Ph.D. (2011) degrees in Telecommunications Engineering from King's College London. He is a Research Assistant Professor with the KIOS Research and Innovation Center of Excellence, University of Cyprus. His interests focus on both basic and applied research on networked intelligent systems. Examples of such systems include intelligent transportation systems, autonomous unmanned aerial systems, and the plethora of cyber-physical systems that arise within IoT.
\end{IEEEbiography}

\begin{IEEEbiography}[{\includegraphics[width=1in,height=1.25in,clip,keepaspectratio]{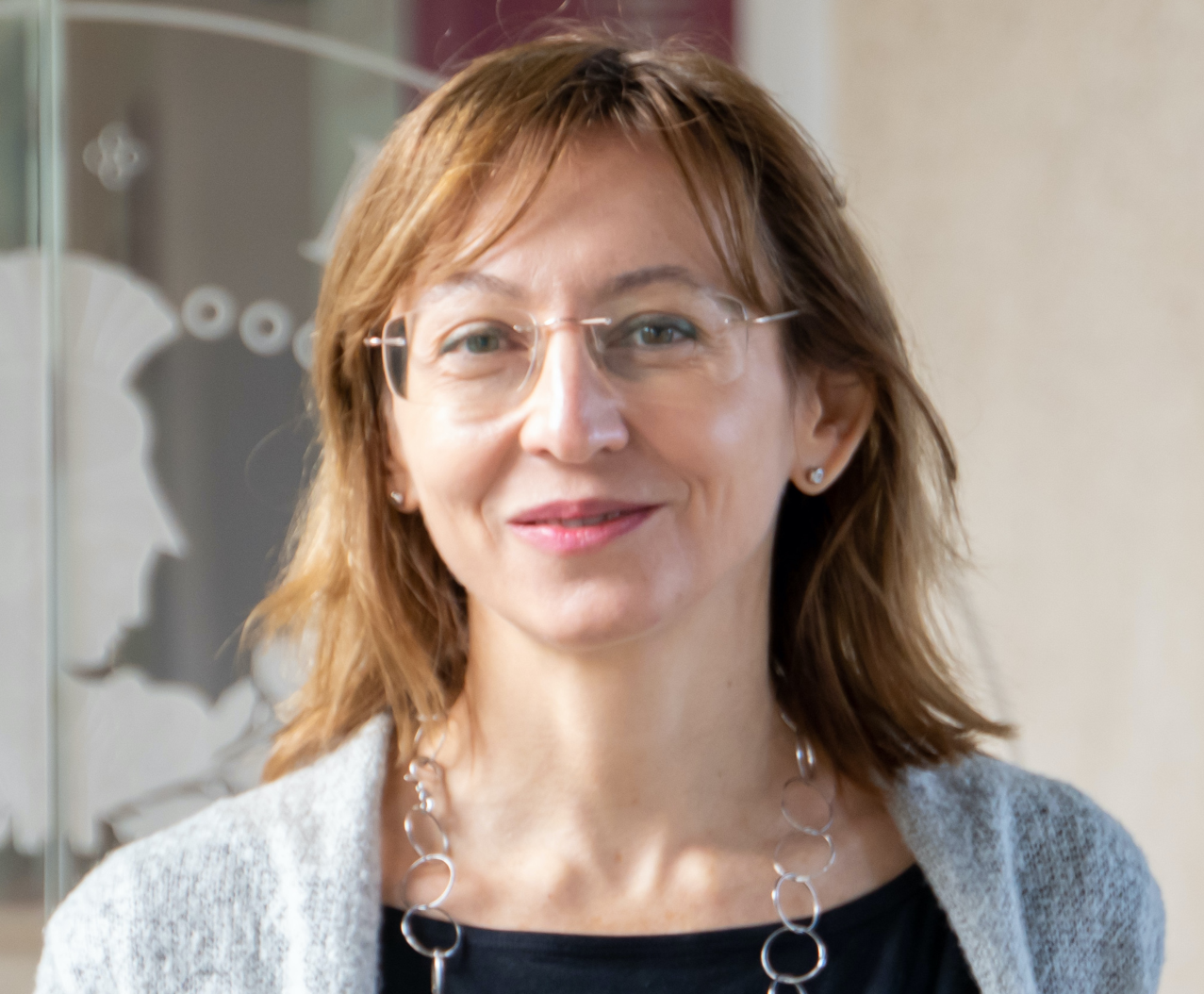}}]{Carla Fabiana Chiasserini} (F'18) worked as a Visiting Researcher at UCSD and as a Visiting Professor at Monash University in 2012 and 2016, and at TUB (Germany) in 2021. She is currently a Professor at Politecnico di Torino and EiC of Computer Communications.
\end{IEEEbiography}

\begin{IEEEbiography}[{\includegraphics[width=1in,height=1.25in,clip,keepaspectratio]{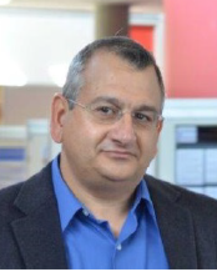}}]{Georgios Ellinas} holds B.Sc., M.Sc., M.Phil., and Ph.D. degrees in Electrical Engineering from Columbia University. He is a Professor at the Department of Electrical and Computer Engineering and a founding member of the KIOS Research and Innovation Center of Excellence at the University of Cyprus. Prior to joining the University of Cyprus, he also served as an Associate Professor of Electrical Engineering at City College of the City University of New York, as a Senior Network Architect at Tellium Inc., and as a Research Scientist/Senior Research Scientist in the Bell Communications Research (Bellcore) Optical Networking Research Group. His research interests are in optical/telecommunication networks, intelligent transportation systems, IoT, and autonomous unmanned aerial systems.
\end{IEEEbiography}

\end{document}